\documentclass[letterpaper, 10pt, conference]{ieeeconf}  

\IEEEoverridecommandlockouts                              

\usepackage[dvipsnames, rgb, table]{xcolor}
\usepackage{lipsum}
\usepackage{soul}
\usepackage{color}
\usepackage[hidelinks]{hyperref}

\usepackage{tikz}
\usepackage{bm}
\usepackage{pgfplots}
\pgfplotsset{compat=1.18}
\usepackage{pgfplotstable}
\usepackage{booktabs}
\usepackage{float}
\usepackage{multirow}
\usepackage{etoolbox}
\usepackage{adjustbox}
\usepackage{diagbox}
\usepackage{graphics} 
\usepackage{epsfig} 
\usepackage{mathptmx} 
\usepackage{times} 
\usepackage{amsmath}
\usepackage{amsfonts}
\usepackage{amssymb} 
\usepackage{multicol}
\usepackage[font=footnotesize]{caption}
\usepackage{threeparttable}
\usepackage{fvextra}  
\DefineVerbatimEnvironment{myverbatim}{Verbatim}
  {breaklines=true, 
  breaksymbolleft={},   
  breakindent=0pt,      
  fontsize=\footnotesize}

\usepgfplotslibrary{fillbetween}
\pgfdeclarelayer{background}
\pgfdeclarelayer{back background}
\pgfsetlayers{back background, background, main}
\usetikzlibrary{calc}
\usetikzlibrary{fit}
\usetikzlibrary{bayesnet}
\usetikzlibrary{arrows}
\usetikzlibrary{arrows.meta}
\usetikzlibrary{positioning}
\usetikzlibrary{shapes}
\usetikzlibrary{shapes.geometric}
\usetikzlibrary{shadows}
\usetikzlibrary{patterns}
\usetikzlibrary{fadings}
\usetikzlibrary{decorations.pathmorphing}
\usetikzlibrary{decorations.pathreplacing}
\usetikzlibrary{decorations.text}

\definecolor{ffblue}{RGB}{097, 108, 140}
\definecolor{ffdarkgreen}{RGB}{086, 140, 135}
\definecolor{fflightgreen}{RGB}{178, 213, 155}
\definecolor{ffyellow}{RGB}{242, 222, 121}
\definecolor{ffred}{RGB}{217, 095, 024}

\definecolor{ffred_pv}{RGB}{202, 074, 046}
\definecolor{fforange_pv}{RGB}{232, 141, 047}
\definecolor{ffgreen_pv}{RGB}{059, 165, 149}
\definecolor{ffgreendark_pv}{RGB}{032, 117, 106}
\definecolor{nature_tab_gray1}{HTML}{D8D6C2}
\definecolor{nature_tab_gray2}{HTML}{ECEADF}
\title{\LARGE \bf
Robotic Long-Horizon  Manipulation with Progressive In-Context Code Generation and Episodic Feedback
}

\author{Yuan Meng$^{1}$, 
Xiangtong Yao$^{1}$,
Haihui Ye$^{1}$,
Yirui Zhou$^{1}$,
Shengqiang Zhang$^{2}$,
Zhenguo Sun$^{3}$,\\
Xukun Li$^{1}$,
Zhenshan Bing$^{4,\dagger}$,
and Alois Knoll$^{1}$ \textit{IEEE fellow}
\thanks{$^{1}$The School of Computation, Information and Technology, Technical University of Munich, Germany.}%
\thanks{$^{2}$The Center for Information and Language Processing, Ludwig Maximilian University of Munich, Germany.}%
\thanks{$^{3}$Beijing Academy of Artificial Intelligence, Bejing, China}
\thanks{$^{4}$The State Key Laboratory for Novel Software Technology, School of Intelligence Science and Technology, Nanjing University, China.}%
\thanks{$^{\dagger}$ Corresponding author: \tt\small bing@nju.edu.cn}
\thanks{Videos and data are available at \url{https://ghiara.github.io/DAHLIA/}}
}

\begin{document}

\maketitle
\thispagestyle{empty}
\pagestyle{empty}

\begin{abstract}
Robotic long-horizon manipulation requires robots to compose perception, reasoning, and action over extended task sequences, yet existing language-conditioned frameworks often rely on dense step-wise feedback, learned action policies, or unstructured prompting, which limits robustness and generalization. We propose DAHLIA, a data-agnostic code-generation framework that treats long-horizon manipulation as episodic task planning and evaluation. DAHLIA uses progressively organized in-context examples with chain-of-thought reasoning to guide an LLM in synthesizing executable code plans from language instructions. To improve robustness under partial observability, a VLM-based reporter evaluates task outcomes only after each execution episode and provides structured feedback for replanning, avoiding the overhead of per-step inference. Experiments on LoHoRavens, CALVIN, Franka Kitchen, and real-world manipulation tasks show that DAHLIA achieves strong performance on over 30 long-horizon tasks and improves generalization to unseen scenarios, including cluttered scenes and occluded objects.
\end{abstract}

\section{INTRODUCTION}
Long-horizon robotic manipulation requires a robot to interpret high-level language instructions, reason over object relations, and execute temporally extended action sequences. Recent vision-language-action models (VLAs) have shown promising progress in language-conditioned manipulation, but they typically require large-scale robot data and extensive training, while their performance often degrades when tasks become longer, more compositional, or substantially different from the training distribution~\cite{yao2023bridging,intelligence2025pi_,meng2025preserving}. This limitation is particularly critical in practical manipulation scenarios, where robots must solve diverse tasks under partial observability, clutter, and changing object configurations.
%
\begin{figure}[t!]
    \centering
    \resizebox{.48\textwidth}{!}{
    \begin{tikzpicture}
        
        \fill[white, rounded corners=4pt, shading=axis, left color=ffgreen_pv!50, right color=fforange_pv!50]
            (-4.5cm,-1cm) -- (4.5cm,-1cm) -- (4.5cm,-2.8cm) -- (-4.5cm,-2.8cm) -- cycle;
        \fill[white] (0,0) -- (4.6cm, 0) -- (4.6cm, -2.9cm) -- (0, -2.9cm) --cycle;
        \fill[white, rounded corners=4pt, shading=axis, left color=ffgreen_pv!50, right color=fforange_pv!50]
            (-4.5cm, 1cm) -- (4.5cm, 1cm) -- (4.5cm, 2.8cm) -- (-4.5cm, 2.8cm) -- cycle;
        \fill[white] (0,0) -- (-4.6cm, 0) -- (-4.6cm, 2.9cm) -- (0, 2.9cm) --cycle;
        \node[inner sep=-1pt, circle, draw=ffgreen_pv!50, fill=white, line width=1pt, label={[label distance=-0.55cm]-90:\scriptsize\textsf{User}}](user)at(-3.5cm,2cm){\includegraphics[width=1cm]{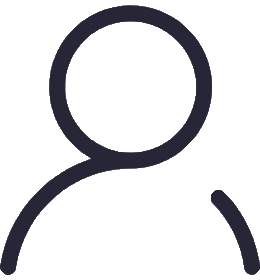}};
        \node[inner sep=-1pt, circle, draw=ffgreen_pv!50, fill=none, line width=1pt, minimum size=1.5cm](line)at(-3.5cm,2cm){};
        \node[inner sep=0pt, circle, draw=none, left color=ffgreen_pv!50, right color=fforange_pv!50, line width=1pt, minimum size=1.525cm](line2)at(3.5cm,-1.8cm){};
        \node[inner sep=-1pt, circle, draw=white, left color=ffgreen_pv!50, right color=fforange_pv!50, line width=1pt, minimum size=1.425cm](line2)at(3.5cm,-1.8cm){};
        \node[inner sep=-1pt, circle, draw=none, fill=white, label={[label distance=0cm]-90:\scriptsize\textsf{Code Plan}}](code)at(3.5cm,-1.8cm){\includegraphics[width=1cm]{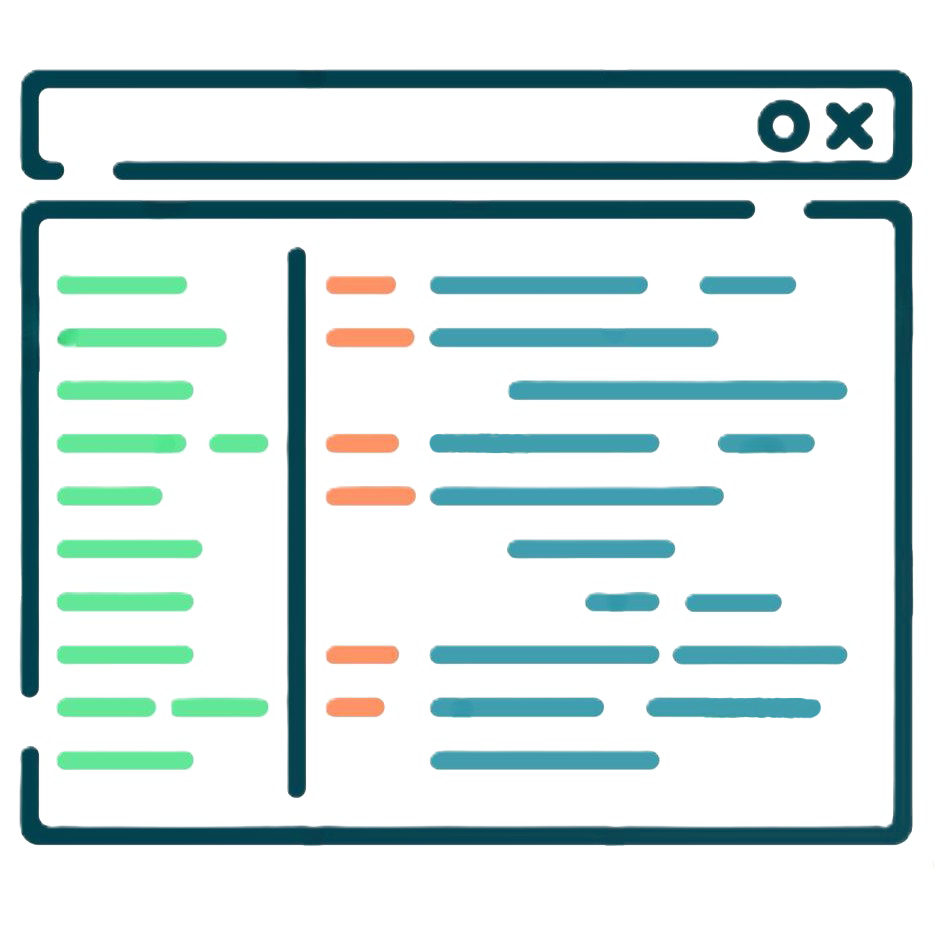}};

        \node[draw=white, fill=ffgreen_pv!40, rectangle, rounded corners=4pt, text width=1.3cm, below left](text)at([yshift=-.3cm]user.south east){\footnotesize\textsf{Build a red head pyramid?}};
        \node[draw=white, fill=fforange_pv!40, rectangle, rounded corners=4pt, text width=1.3cm, above right](text2)at([yshift=.325cm]code.north west){\footnotesize\textsf{Red block is on the top. Succeed.}};
        
        \shade[left color=ffgreen_pv!50, right color=fforange_pv!50] (0,0) circle (2.8cm);
        \node[circle, draw=white, fill=none, line width=4pt, minimum size=4cm](line)at(0,0){};
        \draw[draw=white, line width=3pt, rounded corners=2pt] (0, -1.95cm) -- (0.5cm, -2.4cm) -- (0, -2.85cm);
        \draw[draw=white, line width=3pt, rounded corners=2pt] (0, 1.95cm) -- (-0.5cm, 2.4cm) -- (0, 2.85cm);
        \path[decorate,decoration={text along path, text={|\sf\bfseries\color{white}|Planner Tunnel},text align=center, raise=-0.2cm}] (180:2.5cm) arc (180:120:2.5cm);
        \path[decorate,decoration={text along path, text={|\sf\bfseries\color{white}|Reporter Tunnel},text align=center, raise=-0.2cm}] (0:2.5cm) arc (0:-60:2.5cm);
        \node[]at(2.4, 2.6){\footnotesize \bfseries\textcolor{white}{\textsf{Structured Context Feedback}}};
        \node[]at(2.4, 2.2){\footnotesize \bfseries\textcolor{white}{\textsf{Partial Observability}}};
        \node[]at(2.4, 1.8){\footnotesize \bfseries\textcolor{white}{\textsf{Failure Recovery}}};
        \node[]at(-2.2,-2.2){\footnotesize \bfseries\textcolor{white}{\textsf{Chain-of-Thought}}};
        \node[]at(-2.2,-2.6){\footnotesize \bfseries\textcolor{white}{\textsf{Incremental Few-shot adaptation}}};
        \clip (0,0) circle (2cm) node (scene) {\includegraphics[width=4cm]{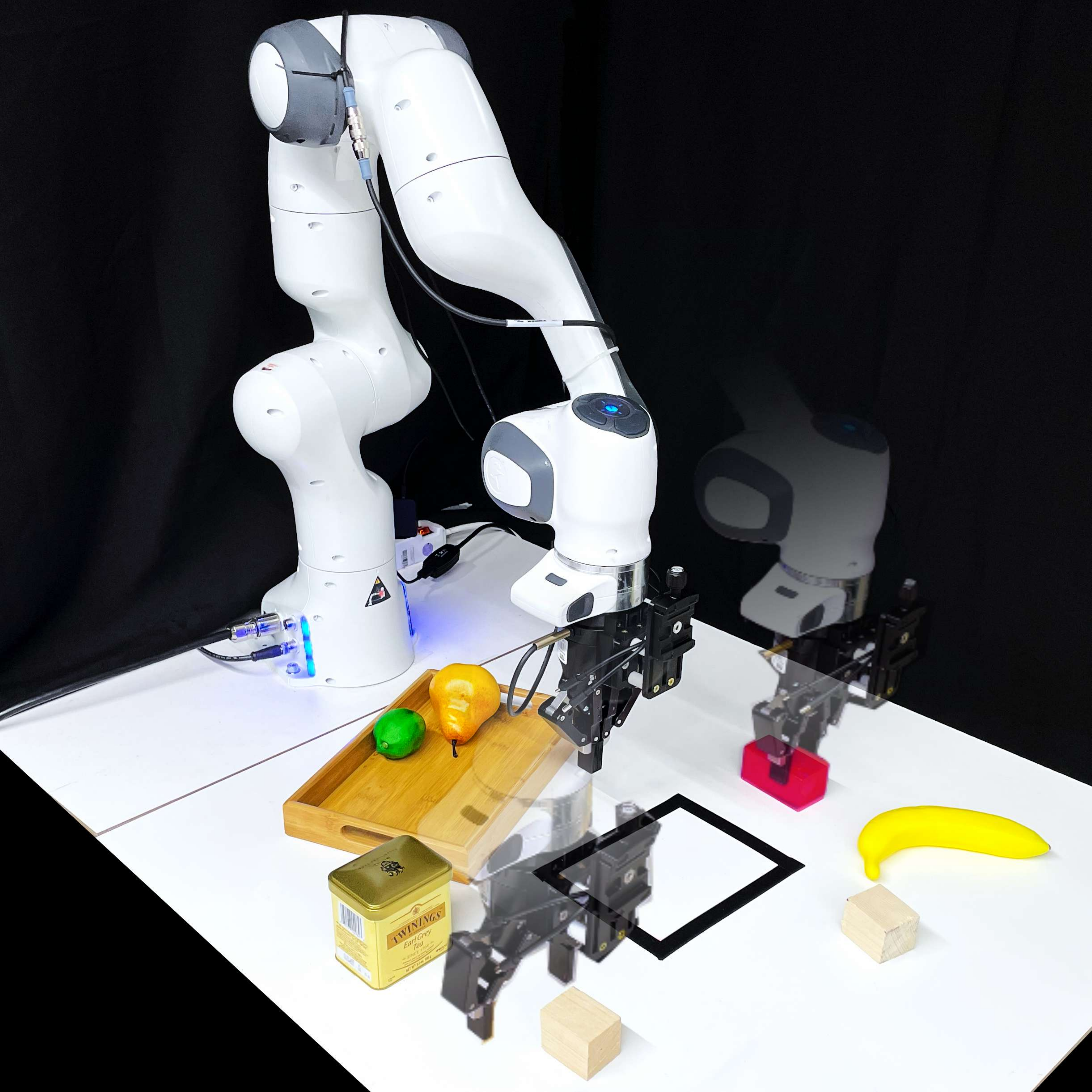}};
        \path[decorate,decoration={text along path, text={|\sf\bfseries\color{white}|Long-Horizon Tasks},text align=center, raise=-0.2cm}] (90:1.75cm) arc (90:-20:1.75cm);

    \end{tikzpicture}
    }
    \vskip -0.05in
    \caption{Overview of our proposed framework. The dual-tunnel structure generates executable task plans and enables task-level closed-loop feedback.}
    \label{fig:Overview}
    \vskip -0.25in
\end{figure}
Large language models provide an alternative route by serving as high-level task planners. Existing hierarchical approaches decompose instructions into subgoals and execute them with pretrained low-level policies or manipulation skills~\cite{ahn2022can,huang2023inner,guo2024doremi,zhang2023lohoravens}. Although such methods improve semantic task decomposition, they often depend on the reliability of the underlying action policy and require frequent feedback or step-wise reasoning. These requirements increase inference cost and make the system vulnerable to cumulative errors, especially in long-horizon tasks where a single incorrect intermediate decision may lead to task failure.

Code-generation-based manipulation has recently emerged as a promising paradigm, where an LLM directly synthesizes executable programs that compose predefined robot APIs~\cite{liang2023code}. Compared with natural-language plans, executable code provides a more structured interface for long-horizon task execution and can reduce dependence on task-specific policy learning. However, existing code-generation methods still face two main challenges. First, many approaches operate in an open-loop manner or use only coarse feedback, making it difficult to recover from failures caused by occlusion, spatial misalignment, or incomplete execution. Second, their few-shot prompts are often arranged without an explicit learning structure, which can cause the model to imitate surface-level code patterns rather than infer reusable task-planning abstractions. As a result, generalization to unseen long-horizon tasks remains limited.

To address these issues, we propose DAHLIA, a \textbf{D}ata-\textbf{A}gnostic \textbf{H}ierarchical Closed-loop Framework for \textbf{L}ong-horizon \textbf{I}ncremental In-context \textbf{A}daptation. DAHLIA formulates manipulation as an episodic code-generation process. In each episode, a planner generates a complete executable task plan by composing predefined perception and manipulation APIs. The plan is executed as a sequence of primitives without per-step replanning. After execution, a reporter evaluates the task outcome from paired observations and the original instruction, then returns structured task-level feedback for the next planning episode if the goal is not yet achieved. This design introduces closed-loop recovery at the task level while avoiding continuous inference during execution.

A central component of DAHLIA is its structured in-context learning strategy. Instead of using randomly selected demonstrations, we organize few-shot examples in progressively increasing functional and structural complexity, and combine them with chain-of-thought reasoning cues. The examples therefore serve as a curriculum that guides the LLM to extract reusable program structures, task decomposition logic, and spatial reasoning patterns. This progressive prompting strategy improves long-horizon code synthesis and supports out-of-distribution generalization without task-specific training.
The main contributions of this work are summarized as follows:
\begin{itemize}
\item We propose an episodic code-generation framework for long-horizon robotic manipulation, enabling an LLM to synthesize executable task plans from language instructions without task-specific training or pretrained low-level action policies.
\item We introduce a progressive in-context learning strategy with chain-of-thought reasoning, which organizes examples from simple to complex and improves compositional generalization in long-horizon planning.
\item We design a VLM-based episodic reporter that provides structured task-level feedback for replanning under partial observability, while avoiding the computational overhead of per-step feedback.
\item We evaluate DAHLIA on more than 30 long-horizon tasks across LoHoRavens, CALVIN, Franka Kitchen, and real-world scenarios, demonstrating improved robustness and generalization over strong language- and code-generation baselines.
\end{itemize}

\section{RELATED WORK}

\textbf{Task planning with language guidance} uses LLMs as high-level planners for task decomposition via natural language, combined with pretrained low-level controllers (e.g., RL-based) to execute long-horizon tasks.
SayCan \cite{ahn2022can} introduced open-loop planning with LLMs, while closed-loop variants like Inner Monologue \cite{huang2023inner}, DoReMi \cite{guo2024doremi}, and LoHoRavens \cite{zhang2023lohoravens} incorporate various feedback mechanisms to enhance robustness.
However, these approaches often struggle in unseen scenarios and degrade under environment perturbations due to the limitations of their low-level policies \cite{zhang2023lohoravens,shridhar2022cliport}.
While feedback has been well studied in language-generation-based frameworks, its potential remains underexplored in the context of code-generation-based planning.

\textbf{Manipulation with code generation} leverages LLMs to directly control robots through code, bypassing the need for fine-tuned low-level policies \cite{chen2025lemmo,zhang2023lohoravens,liang2023code}. 
Recent works fall into categories such as direct code generation \cite{liang2023code,chen2024roboscript,arenas2024prompt}, task decomposition \cite{mu2024robocodex}, and constraint-based planning \cite{huang2023voxposer,huang2025rekep}, with Code-as-Policy (CaP) \cite{liang2023code} being a notable example.
However, few studies explore both the role of structured few-shot prompting on robustness and generalization, and closed-loop feedback. 
Our work aims to address this gap by incorporating a closed-loop feedback pipeline, combining examples with progressively increasing functionality and difficulty.

\begin{figure*}[ht!]
    \centering
    \resizebox{\textwidth}{!}{
    \begin{tikzpicture}
        \shade[left color=ffgreendark_pv!40, right color=fforange_pv!40] (15,-5) circle (5.75cm);
        \node[circle, fill=white, minimum size=10cm] at (15,-5){};
        \fill[ffgreendark_pv!50, rounded corners=4pt, shading=axis, left color=ffgreendark_pv!50, right color=ffgreendark_pv!20]
            (5,0) -- (10,0) -- (15,-5) -- (10,-10) -- (5, -10) -- cycle;
        \fill[fforange_pv!50, rounded corners=4pt, shading=axis, left color=fforange_pv!20, right color=fforange_pv!50]
            (15, -5) -- (20,0) -- (25,0) -- (25,-10) -- (20,-10) -- cycle;
        \node[circle, draw=none, fill=white, line width=1pt, minimum size=9cm](white_circle)at(15,-5){};
        \shade[left color=ffgreen_pv!50, right color=fforange_pv!50] (15,-5) circle (4.25cm);
        \node[circle, fill=white, minimum size=8cm] (env_bg) at (15,-5){};
        
        \node[circle, fill=none, minimum size=1cm](env)at(15,-5){\includegraphics[width=8cm]{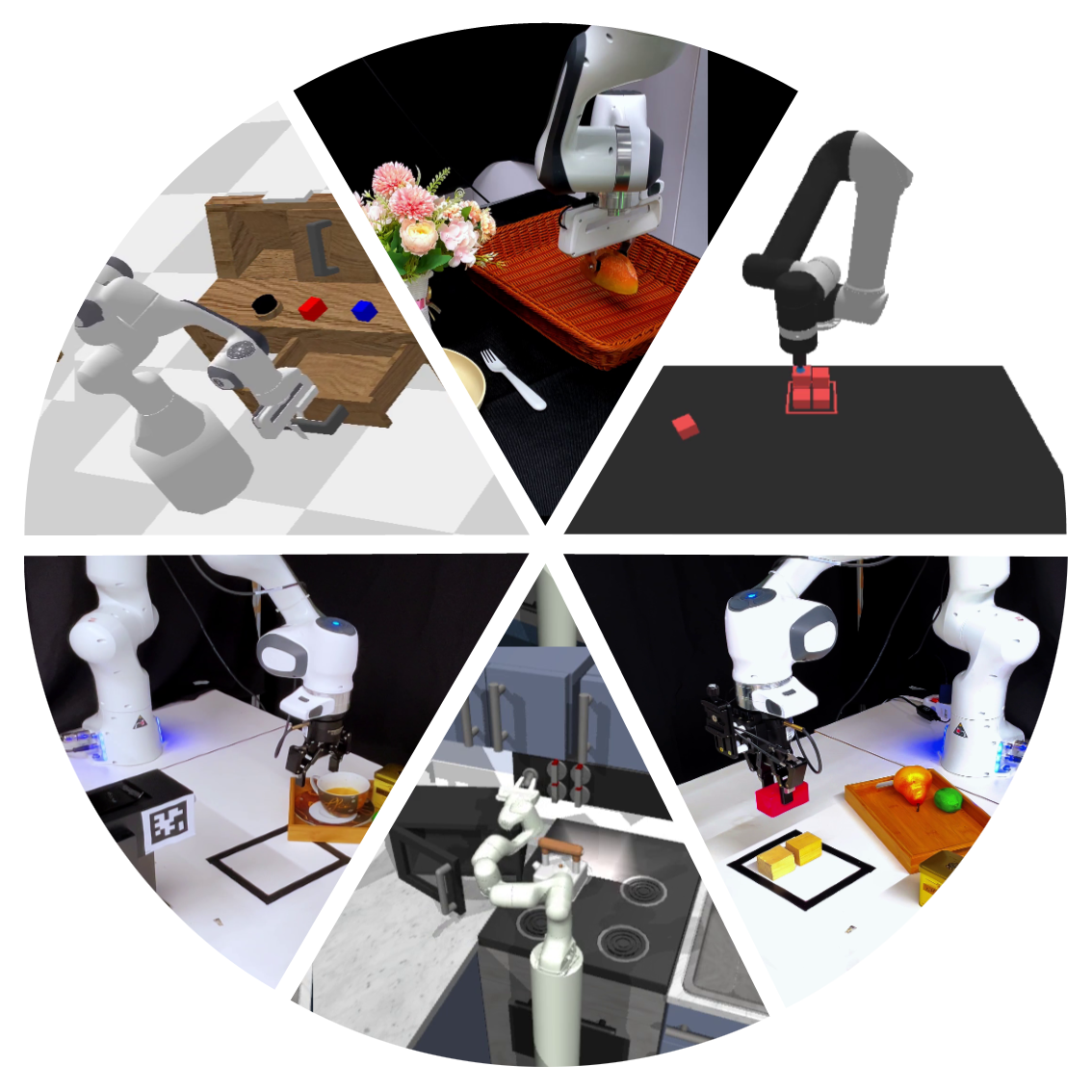}};
        \node[circle, fill=white] (tasks) at (15,-5){\textbf{\textsf{Tasks}}};

        \fill[fflightgreen!60, rounded corners=3pt, shading=axis, left color=fflightgreen!20, right color=fflightgreen!60]
            (-5,0) -- (4,0) -- (4.5,-0.5) -- (4.5,-1.5) -- (-5, -1.5) -- cycle;
        \fill[fflightgreen!20, rounded corners=3pt]
            (-4.75,-2.25) -- (4.5,-2.25) -- (4.5,-10) -- (-4.75, -10) -- cycle;
        \node[inner sep=-1pt, circle, draw=fflightgreen, fill=white, line width=1pt, label={[label distance=-0.55cm]-90:\small \textsf{User}}](user)at(-5cm,-1cm){\includegraphics[width=1.5cm]{imgs/user.png}};
        \node[inner sep=-1pt, circle, draw=fflightgreen, fill=none, line width=1pt, minimum size=2.3cm](line)at(-5cm,-1cm){};

        \node[text width=7.5cm, right]at([xshift=.25cm, yshift=.2cm]user.east){\large \textsf{Can you use blocks to build a 2*2*2 cube in the zone?}};
        \node[]at([xshift=2cm, yshift=-.1cm]user.south east){\large \textcolor{fflightgreen}{\textbf{\textsf{Prompted with:}}}};
        
        \node[rectangle, draw=none, fill=fflightgreen!60, rounded corners=3pt, minimum width=8.5cm, minimum height=1.25cm, text width=8cm, align=left](prompt1)at([xshift=4.85cm, yshift=-1.25cm]user.south){\large \textbf{\textsf{1. Available APIs:}} \\ \small \quad\textsf{from utils import get\_obj\_names, get\_bbox, ...}};
        \node[rectangle, draw=none, fill=fflightgreen!60, rounded corners=3pt, minimum width=8.5cm, minimum height=1.25cm, text width=8cm, align=left](prompt2)at([yshift=-1cm]prompt1.south){\large \textbf{\textsf{2. Coordinates:}} \\ \small \quad\textsf{front:x+, left:y+, top:z+, back:x-, right:y-, below:z-}};
        \node[rectangle, draw=none, fill=fflightgreen!60, rounded corners=3pt, minimum width=8.5cm, minimum height=1.25cm, text width=8cm, align=left](prompt3)at([yshift=-1cm]prompt2.south){\large \textbf{\textsf{3. Rules:}} \\ \small \quad\textsf{You are writing Python code for robot manipulation ...}};
        \node[rectangle, draw=none, fill=fflightgreen!60, rounded corners=3pt, minimum width=8.5cm, minimum height=1.25cm, text width=8cm, align=left](prompt4)at([yshift=-1cm]prompt3.south){\large \textbf{\textsf{4. Example plan (increasing difficulty):}} \\ \small \quad\textsf{def stack\_blocks(...); def stack\_blocks\_by\_color(...) ...}};
        \node[]at([yshift=-0.5cm]prompt4.south){\large \textcolor{fflightgreen}{\textbf{\textsf{Chain-of-Thought Reasoning}}}};

        \draw[-{Circle}, draw=fflightgreen!80, line width=3pt, rounded corners=1pt] (user.south) |- ([xshift=7pt]prompt1.west);
        \draw[-{Circle}, draw=fflightgreen!80, line width=3pt, rounded corners=1pt] (user.south) |- ([xshift=7pt]prompt2.west);
        \draw[-{Circle}, draw=fflightgreen!80, line width=3pt, rounded corners=1pt] (user.south) |- ([xshift=7pt]prompt3.west);
        \draw[-{Circle}, draw=fflightgreen!80, line width=3pt, rounded corners=1pt] (user.south) |- ([xshift=7pt]prompt4.west);

        \node[rectangle, draw=none, fill=ffdarkgreen!40, rounded corners=3pt, minimum width=4cm, minimum height=1.25cm](planner_bg)at([xshift=11.7cm, yshift=-1.2cm]user.east){};
        \draw[->, draw=ffdarkgreen, line width=2pt, rounded corners=1pt] ([xshift=-0.2cm]planner_bg.south east) to[out=-60, in=60] ++ (0, -2.4cm);
        \node[rectangle, draw=none, fill=ffdarkgreen!30, rounded corners=3pt, minimum width=4cm, minimum height=1.25cm](planner)at([xshift=11.5cm, yshift=-1cm]user.east){\Large \textsf{Planners}};
        \node[rectangle, draw=none, fill=ffdarkgreen!30, rounded corners=3pt, minimum width=4cm, minimum height=1.25cm](tasks)at([yshift=-1.3cm]planner.south){\Large \textsf{Task plan}};
        \node[rectangle, draw=none, fill=ffdarkgreen!30, rounded corners=3pt, minimum width=4cm, minimum height=1.25cm](lmp)at([yshift=-1.3cm]tasks.south){\Large \textsf{LMPs/APIs}};
        \node[rectangle, draw=none, fill=ffdarkgreen!50, rounded corners=3pt, minimum width=4cm, minimum height=1.25cm](primitives_bg0)at([xshift=0.4cm, yshift=-1.7cm]lmp.south){};
        \node[rectangle, draw=none, fill=ffdarkgreen!40, rounded corners=3pt, minimum width=4cm, minimum height=1.25cm](primitives_bg1)at([xshift=0.2cm, yshift=-1.5cm]lmp.south){};
        \node[rectangle, draw=none, fill=ffdarkgreen!30, rounded corners=3pt, minimum width=4cm, minimum height=1.25cm](primitives)at([yshift=-1.3cm]lmp.south){\Large \textsf{Primitives}};
        \node[](dots)at([yshift=-0.5cm]primitives_bg1.south){\Large \textsf{\textbf{\dots}}};

        \draw[->, draw=ffdarkgreen, line width=2pt, rounded corners=1pt] (planner.south) -- (tasks.north);
        \draw[->, draw=ffdarkgreen, line width=2pt, rounded corners=1pt] (tasks.south) -- (lmp.north);
        \draw[->, draw=ffdarkgreen, line width=2pt, rounded corners=1pt] (lmp.south) -- (primitives.north);

        \node[rectangle, draw=none, fill=fforange_pv!20, rounded corners=3pt, minimum width=4cm, minimum height=1.25cm](reporter)at([xshift=26.5cm, yshift=-1cm]user.east){\Large \textsf{Reporter}};
        \node[rectangle, draw=none, fill=fforange_pv!20, rounded corners=3pt, minimum width=4cm, minimum height=1.25cm](eval)at([yshift=-1.3cm]reporter.south){\Large \textsf{Evaluation}};
        \node[rectangle, draw=none, fill=fforange_pv!20, rounded corners=3pt, minimum width=4cm, minimum height=1.25cm](lmp2)at([yshift=-1.3cm]eval.south){\Large \textsf{LMPs/APIs}};
        \node[rectangle, draw=none, fill=fforange_pv!20, rounded corners=3pt, minimum width=4cm, minimum height=1.25cm](obs)at([yshift=-1.3cm]lmp2.south){\Large \textsf{Observation}};

        \draw[->, draw=fforange_pv, line width=2pt, rounded corners=1pt] (obs.north) -- (lmp2.south);
        \draw[->, draw=fforange_pv, line width=2pt, rounded corners=1pt] (lmp2.north) -- (eval.south);
        \draw[->, draw=fforange_pv, line width=2pt, rounded corners=1pt] (eval.north) -- (reporter.south);

        \fill[draw=none, fill=ffyellow!15, rounded corners=4pt,]
            (25.5,0) -- (38.5, 0) -- (38.5,-10) -- (25.5,-10) -- cycle;
        \node[draw=none, fill=none]at(35,-.5){\large \textsf{\textbf{\textcolor{black}{Structured Contextual Feedback:}}}};
        \node[rectangle, draw=none, fill=ffyellow!70, rounded corners=3pt, minimum width=6.5cm, minimum height=1.25cm, text width=6cm, align=left, right](feedback1)at([xshift=7.2cm]reporter.east){\large \textbf{\textsf{Success signal}} \\ \small \quad\textsf{Boolean: task complete/ not complete.}};
        \node[rectangle, draw=none, fill=ffyellow!70, rounded corners=3pt, minimum width=6.5cm, minimum height=1.25cm, text width=6cm, align=left,](feedback2)at([yshift=-1cm]feedback1.south){\large \textbf{\textsf{Object}} \\ \small \quad\textsf{Str: which [object] to operate.}};
        \node[rectangle, draw=none, fill=ffyellow!70, rounded corners=3pt, minimum width=6.5cm, minimum height=1.25cm, text width=6cm, align=left,](feedback3)at([yshift=-1cm]feedback2.south){\large \textbf{\textsf{Location}} \\ \small \quad\textsf{Array: where the objects [locations].}};
        \node[rectangle, draw=none, fill=ffyellow!70, rounded corners=3pt, minimum width=6.5cm, minimum height=1.25cm, text width=6cm, align=left,](feedback4)at([yshift=-1cm]feedback3.south){\large \textbf{\textsf{Target}} \\ \small \quad\textsf{[Str, Array]: where the [target] located.}};
        \node[rectangle, draw=none, fill=ffyellow!70, rounded corners=3pt, minimum width=6.5cm, minimum height=1.25cm, text width=6cm, align=left,](feedback5)at([yshift=-1cm]feedback4.south){\large \textbf{\textsf{Action}} \\ \small \quad\textsf{Str: what [action] primitive to take}};

        \node[label={[label distance=-0.2cm]-90:\large \textbf{\textsf{Before}}}, right](state1)at([xshift=1.1cm, yshift=-.3cm]obs.east){\includegraphics[width=2.5cm]{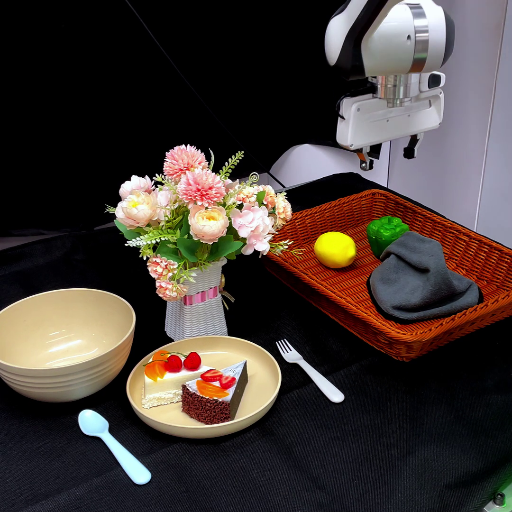}};
        \node[label={[label distance=-0.2cm]-90:\large \textbf{\textsf{After}}}, right](state2)at(state1.east){\includegraphics[width=2.5cm]{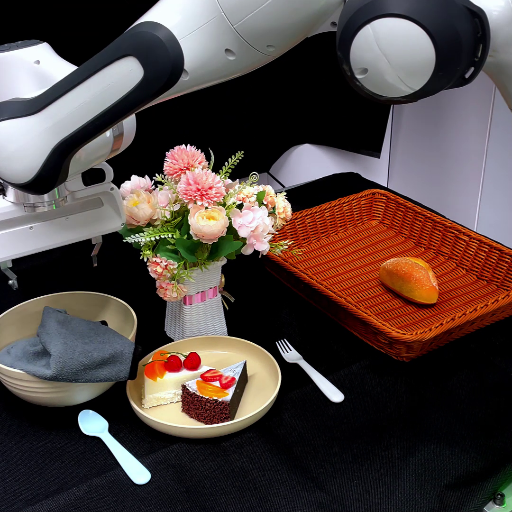}};
        \node[above right]at([yshift=-.1cm]state1.north west){\large \textbf{\textsf{Loop 1}}};
        \node[rectangle, draw=none, fill=ffyellow!70, rounded corners=3pt, minimum width=5cm, minimum height=1.25cm, text width=4.75cm, align=left, left](fe)at([xshift=-.75cm]feedback3.west){\textsf{A bread at (x1, y1, z1) is not yet placed in the bowl. Task not complete.}};
        \node[above]at(fe.north){\Large\textbf{...}};
        \node[label={[label distance=-0.2cm]-90:\large \textbf{\textsf{Before}}}](state3)at([yshift=4.5cm]state1.north){\includegraphics[width=2.5cm]{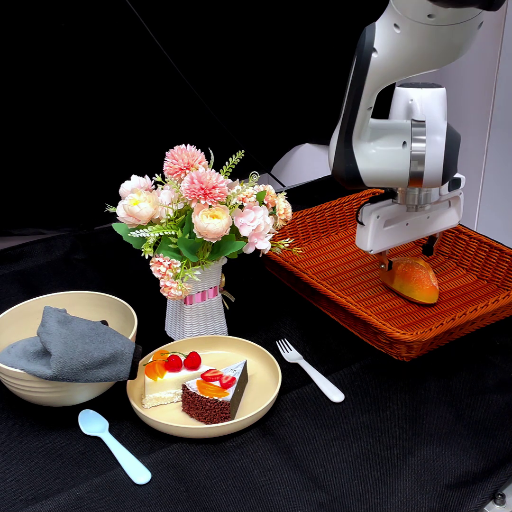}};
        \node[label={[label distance=-0.2cm]-90:\large \textbf{\textsf{After}}}, right](state4)at(state3.east){\includegraphics[width=2.5cm]{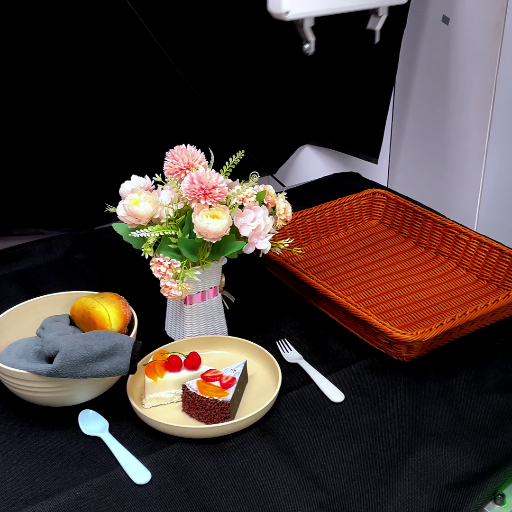}};
        \node[above right]at([yshift=-.1cm]state3.north west){\large \textbf{\textsf{Loop k}}};
        \draw[-{Circle}, draw=ffyellow!90, line width=3pt, rounded corners=1pt] (fe.east) -| ([xshift=-.4cm]feedback1.west) -- ([xshift=7pt]feedback1.west);
        \draw[-{Circle}, draw=ffyellow!90, line width=3pt, rounded corners=1pt] (fe.east) -| ([xshift=-.4cm]feedback2.west) -- ([xshift=7pt]feedback2.west);
        \draw[-{Circle}, draw=ffyellow!90, line width=3pt, rounded corners=1pt] (fe.east) -| ([xshift=-.4cm]feedback3.west) -- ([xshift=7pt]feedback3.west);
        \draw[-{Circle}, draw=ffyellow!90, line width=3pt, rounded corners=1pt] (fe.east) -| ([xshift=-.4cm]feedback4.west) -- ([xshift=7pt]feedback4.west);
        \draw[-{Circle}, draw=ffyellow!90, line width=3pt, rounded corners=1pt] (fe.east) -| ([xshift=-.4cm]feedback5.west) -- ([xshift=7pt]feedback5.west);
        \draw[-{Triangle Cap []. Fast Triangle[]}, draw=fforange_pv!80, line width=8pt, rounded corners=1pt] (obs.east) --++ (1.25cm, 0);
        \draw[-{Triangle Cap []. Fast Triangle[]}, draw=fforange_pv!80, line width=8pt, rounded corners=1pt] (fe.west) -| ([xshift=.75cm]reporter.east) -- (reporter.east);
        

        \begin{scope}[shift={(15,-5)}] 
            \path[decorate, decoration={text along path, text={|\sf|All items cleared from basket, task complete.}, text align=center}] (135:5.25) arc (135:45:5.25);
            \path[decorate,decoration={text along path, text={|\sf|Is task ``clear basket'' succeed?}, text align=center}] (225:5.5) arc (225:315:5.5);
        \end{scope}

        \draw[-{Triangle Cap []. Fast Triangle[]}, draw=fflightgreen!80, line width=8pt, rounded corners=1pt] ([xshift=-1cm]planner.west) -- (planner.west);
        \draw[-{Triangle Cap []. Fast Triangle[]}, draw=ffdarkgreen!80, line width=8pt, rounded corners=1pt] ([xshift=1.7cm, yshift=0.8cm]planner.east) to[out=225, in=0] (planner.east);
        \draw[-{Triangle Cap []. Fast Triangle[]}, draw=ffdarkgreen!80, line width=8pt, rounded corners=1pt] (env_bg) to[out=143, in=0] (planner.east);
        \draw[-{Triangle Cap []. Fast Triangle[]}, draw=ffdarkgreen!80, line width=8pt, rounded corners=1pt] (primitives.east) to[out=0, in=210] (env_bg);
        \draw[-{Triangle Cap []. Fast Triangle[]}, draw=ffdarkgreen!80, line width=8pt, rounded corners=1pt] (primitives.east) to[out=0, in=135] ([xshift=1.75cm, yshift=-1cm]primitives.east);
        \draw[-{Triangle Cap []. Fast Triangle[]}, draw=fforange_pv!80, line width=8pt, rounded corners=1pt] (env_bg) to[out=325, in=180] (obs.west);
        \draw[-{Triangle Cap []. Fast Triangle[]}, draw=fforange_pv!80, line width=8pt, rounded corners=1pt] ([xshift=-1.75cm, yshift=-1cm]obs.west) to[out=45, in=180] (obs.west);
        \draw[-{Triangle Cap []. Fast Triangle[]}, draw=fforange_pv!80, line width=8pt, rounded corners=1pt] (reporter.west) to[out=180, in=-45] ([xshift=-1.7cm, yshift=0.8cm]reporter.west);
        
        \node[draw=none, fill=none]at(6.5,-0.5){\LARGE \textsf{\textbf{\textcolor{white}{PLANNER}}}};
        \node[draw=none, fill=none]at(23.25,-0.5){\LARGE \textsf{\textbf{\textcolor{white}{REPORTER}}}};
    \end{tikzpicture}
    }
    \vskip -0.05in
    \caption{Framework of DAHLIA. Our framework utilizes a dual-tunnel pipeline to implement a closed-loop feedback at the task level for various long-horizon manipulation tasks. The planner, powered by LLMs, converts human instructions into executable code plans, leveraging few-shot incremental adaptation in CoT to generate and perform multi-step primitives at once. The reporter, powered by a VLM, evaluates task outcomes based on paired visual observations and task instructions, then provides structured contextual feedback to the planner, enabling re-planning and robust performance in unstructured environments.}
    \vskip -0.2in
    \label{fig:framework}
\end{figure*}

\section{METHOD}
\subsection{Problem setting}\label{sec:method_setting}
We consider embodied long-horizon manipulation at the task-planning level, where a robot interprets a natural language instruction and executes a sequence of actions to achieve a semantic goal. Our focus is on the generation, evaluation, and revision of task-level executable plans, rather than continuous control or low-level policy learning.

A \textbf{task} is specified by a language instruction $l$ and an initial environment state $s_0$ obtained through perception. The objective is to generate an executable code plan composed of mid-level manipulation primitives that transforms the system into a goal-satisfying terminal state. We assume access to a library of predefined Application Programming Interfaces (APIs) (e.g., pick-and-place, pose queries), which encapsulate temporal constant low-level motion execution with fixed interfaces.

We adopt an \textbf{episodic execution model with temporal abstraction} to describe the process logic of our framework. In each episode, the system (i) generates a complete task plan as executable code, (ii) executes the plan open-loop as a sequence of primitives, and (iii) evaluates the outcome after the entire plan finishes. No per-step or per-skill feedback is used within an episode; feedback is introduced only \textbf{between episodes}.
This design is motivated by the observation that, under code-based execution, primitive actions are typically executed in a predictable and repeatable manner via deterministic APIs, while task failures more often arise from environment-level perturbations such as occlusions that can only be meaningfully assessed after execution. Moreover, avoiding per-step feedback reduces redundant inference overhead, which reduces the efficiency of real-time robotic operation as observed in prior work \cite{belkhale2024rt,guo2024doremi,zhang2023lohoravens}.

Based on this execution model, the system operates in a \textbf{task-level feedback loop}. After each episode, the task outcome is assessed with respect to the original instruction. If the task is incomplete or misaligned, structured task-level feedback is provided to guide the generation of a revised plan in the next episode. This feedback does not modify low-level control, but supports episodic task recovery and replanning under partial observability, distinguishing our approach from closed-loop control or policy-learning-based methods.

\subsection{System interfaces and language conditioning}\label{sec:prompting}
Under the above task-level execution model, the key challenge is to ensure that a language model produces reliable and executable task plans that are consistent with the robot's available capabilities and the environment's spatial conventions. 
To this end, we instantiate the system by explicitly specifying 
(i) the manipulation and sensing interfaces, 
(ii) a shared spatial reference, 
(iii) execution constraints that enforce stable code generation, and 
(iv) progressively organized in-context plan templates that provide structured examples for long-horizon reasoning (Fig.~\ref{fig:framework}, left).

\subsubsection{Manipulation and state interfaces}
We provide the framework with a library of predefined manipulation and scene-query APIs that serve as the only executable building blocks for task plans. 
These interfaces encapsulate low-level motion execution (e.g., \texttt{put\_first\_on\_second}) as well as state access (e.g., object identities, poses, bounding boxes, and free-space queries). 
A generated plan is considered valid only if it calls these APIs with correct argument types and returns code in the prescribed format. 
Table~\ref{tab:api} summarizes the available APIs and their intended functionality.
\begin{table}[ht!]
    \centering
    \caption{Instructions of APIs.}
    \vskip -0.05in
    \resizebox{.49\textwidth}{!}{
    \begin{tabular}{l p{6.7cm}}
        \toprule
        \rowcolor{nature_tab_gray1}
        \textbf{API Name} & \textbf{Instruction} \\
        \midrule
        get\_obj\_names() & Retrieves all objects in the environment with attributes such as color, shape, and unique IDs. \\
        \midrule
        \rowcolor{nature_tab_gray2}
        get\_obj\_pos(obj), get\_obj\_rot(obj) & Obtains the 3D position and rotation (Euler angle or quaternion) of a specific object (\textbf{obj}). \\
        \midrule
        get\_bbox(obj) & Provides the axis-aligned bounding box of an object (\textbf{obj}) to aid in spatial inference. \\
        \midrule
        \rowcolor{nature_tab_gray2}
        denormalize(pos) & Converts normalized coordinates (\textbf{pos}) into actual environment coordinates via linear transformation. \\
        \midrule
        is\_target\_occupied(targ) & Checks if a target position or object (\textbf{targ}) is occupied and returns the names of occupying objects. \\
        \midrule
        \rowcolor{nature_tab_gray2}
        get\_random\_free\_pos(targ, area) & Returns a random free position within a specified area (\textbf{area}) that does not occupy the target (\textbf{targ}). \\
        \midrule
        put\_first\_on\_second(obj1, obj2) & Executes a pick-and-place primitive to position (\textbf{obj1}) on (\textbf{obj2}) or a specified pose. \\
        \bottomrule
    \end{tabular}
    }
    \label{tab:api}
\vskip -0.1in
\end{table}

\subsubsection{Spatial reference and coordinate system}
Long-horizon tasks often require spatial reasoning over relative descriptions (e.g., ``left of'', ``in front of'', ``on top of''), while the underlying environments differ in their coordinate conventions. 
To standardize spatial grounding across tasks, we provide an explicit scene reference frame and axis semantics (e.g., ``front: $x+$'', ``left: $y-$'', ``top: $z+$''). 
The framework can perform intermediate reasoning in a normalized coordinate system and convert to environment coordinates using \texttt{denormalize(pos)} when issuing primitive calls.

\subsubsection{Execution rules and constraints}
To improve generation stability and prevent invalid code, we supply a small set of explicit execution constraints as natural-language rules embedded as comments. 
These constraints specify the role of each component (e.g., writing executable Python plans), restrict the action space to the provided APIs (e.g., no external imports), and encourage safe resolution of ambiguous references via dedicated parsing utilities (e.g., \texttt{parse\_obj\_name()} and \texttt{parse\_position()}).
The following snippet illustrates the type of constraints used:

\begin{myverbatim}
You are writing Python code for robot manipulation. Refer to the code style in the examples below. You can use the existing APIs above; you must NOT import other packages...
\end{myverbatim}

\subsubsection{In-context plan examples with progressive complexity}
Beyond interface specifications, we condition the framework with a small set of example plans to provide templates for composing primitives in long-horizon tasks. 
Naively appending randomly selected examples is often insufficient when the target task differs substantially from the demonstrations. 
We therefore organize examples in a progressively increasing order of functional and structural complexity, and pair them with CoT-style reasoning cues to emphasize \textbf{decomposition and plan logic} rather than surface-form imitation.
Concretely, earlier examples demonstrate basic compositions (e.g., \texttt{stack\_blocks(objs, pos)}), while later ones introduce richer constraints (e.g., color-conditioned stacking via \texttt{stack\_blocks\_by\_color(objs, pos, target\_color)}), and may further include alignment requirements such as rotation-aware placements.
This structured conditioning encourages the framework to reuse reusable program structures and improves robustness when scaling to longer horizons and novel task configurations.

With the above interfaces and contextual conditioning in place, DAHLIA instantiates a dual-tunnel task-level loop consisting of (i) a planner that generates executable code plans and (ii) a reporter that evaluates outcomes and triggers plan revision (Fig.~\ref{fig:framework}). 
Inspired by previous work~\cite{liang2023code}, we treat the language model as a program synthesizer that composes predefined primitives into a coherent task plan.
For reproducibility, we provide more prompting detailed information in Appendix A.

\subsection{Task-level planner}
The planner (Fig.~\ref{fig:framework} darkgreen part) operates at the task level and is responsible for synthesizing executable code plans that compose manipulation primitives to achieve the given instruction. 
Given the system interfaces defined in Sec.~\ref {sec:prompting}, the planner generates a complete task plan as Python code, where each plan corresponds to a single execution episode.
The planner is instantiated as a primary LLM agent, supported by a set of auxiliary language model programs (LMPs) that assist with localized reasoning. 
With direct access to structured scene information through APIs, the primary planner does not rely on end-to-end multimodal inference. 
Instead, it focuses on high-level task decomposition and plan composition, mitigating the accumulation of reasoning errors commonly observed in previous per-step decision pipelines.

Natural language task instructions often involve complex relational or attribute-based references, such as ``the third block from the left that is different in color from the rightmost block but has the same size.''
While the primary planner is responsible for generating the overall task structure, resolving such detailed references can be non-trivial. 
To address this, the planner delegates specific sub-problems to dedicated LMPs, such as parsing object descriptions, grounding spatial expressions, or synthesizing auxiliary functions.
Each LMP performs a narrowly scoped transformation (e.g., parsing object names or positions) and returns structured results to the primary planner. 
This modular delegation allows the planner to reason compositionally without entangling global task logic with low-level parsing details.
The set of LMPs used in our framework is summarized in Table~\ref{tab:lmp}. 
Together, these components enable the planner to construct coherent long-horizon task plans by composing reusable primitives and intermediate reasoning utilities.
\begin{table}[t!]
\vskip 0.06in
    \centering
    \caption{Instructions of LMPs.}
    \vskip -0.05in
    \resizebox{.49\textwidth}{!}{
    \begin{tabular}{l p{6.7cm}}
        \toprule
        \rowcolor{nature_tab_gray1}
        \textbf{LMP Name} & \textbf{Instruction} \\
        \midrule
        parse\_obj\_name(dsc, ctxt) & Filters and returns object names from the context (\textbf{ctxt}) that match the target description (\textbf{dsc}) provided by the planner. \\
        \midrule
        \rowcolor{nature_tab_gray2}
        parse\_position(dsc) & Converts a language description (\textbf{dsc}) of a location into one or more position coordinates or poses. \\
        \midrule
        parse\_function(dsc) & Implements new APIs by parsing the planner's description (\textbf{dsc}) and generating their functionality automatically. \\
        \midrule
        \rowcolor{nature_tab_gray2}
        parse\_completion(dsc, ctxt) & Evaluates task completion from the context (\textbf{ctxt}) that match the target description (\textbf{dsc}). \\
        \bottomrule
    \end{tabular}
    }
    \label{tab:lmp}
\vskip -0.2in
\end{table}

Finally, following the execution model introduced in Sec.~\ref{sec:method_setting}, the planner produces plans that are executed open-loop within an episode. 
All primitives in a plan are executed sequentially without intermediate replanning or per-step feedback. 
This episodic planning strategy leverages temporal abstraction to reduce computational overhead and avoids excessive sensitivity to transient execution noise, while still allowing corrections through inter-episode replanning.

\subsection{Episodic feedback with reporter}
While the planner generates and executes task plans, it does not directly observe whether the resulting environment state satisfies the task goal, particularly under partial observability or execution imperfections.
To close the task-level feedback loop, we introduce a reporter (Fig.~\ref{fig:framework} orange part) that evaluates episodic outcomes after each execution episode.
The reporter is implemented as a VLM whose role is strictly limited to assessing task outcomes rather than generating actions or modifying control.
Operating only at episode boundaries, the reporter evaluates the post-execution state with respect to the original task instruction and determines whether the goal has been achieved.

Concretely, the reporter receives (i) the task instruction, (ii) the current execution task plan generated by the planner, and (iii) paired RGB-D observations captured before and after the execution episode.
Using these inputs, it evaluates task completion via the \texttt{parse\_completion} interface and produces structured task-level feedback.
As explained in Sec.~\ref{sec:method_setting}, unlike step-wise monitoring approaches, the reporter does not intervene during primitive execution and does not reason over intermediate control signals.

The reporter output consists of two components:
(i) a binary success flag indicating whether the task is complete, and 
(ii) a structured description of the residual task state when completion fails.
This structured feedback encodes four key elements: the relevant object, its current location, the intended target, and the missing or incorrect action required to resolve the mismatch.
For example, the reporter may return: 
\textit{“a red block [object] at $(x,y,z)$ [location] is not yet stacked [action] on the target block [target].”}
This explicit, task-oriented representation enables the planner to revise its next plan without relying on implicit or free-form language feedback.
By constraining the reporter output to a structured schema, we reduce ambiguity in failure diagnosis and facilitate consistent plan revision across episodes.

Importantly, we do not require the reporter to have perfect perception, nor does it guarantee correctness in all cases.
Its judgments may be affected by occlusions or perceptual uncertainty, and incorrect assessments can lead to redundant replanning.
Nevertheless, within our episodic execution model, such errors manifest as additional planning iterations rather than unsafe control behavior.
Empirically, we observe that this evaluation-driven replanning mechanism improves robustness in long-horizon tasks where partial observability and delayed failure detection are common.

The overall planner–reporter interaction iterates until the reporter confirms task completion, forming a task-level episodic feedback loop that supports recovery from execution errors without requiring continuous control feedback.

\begin{figure*}[ht!]
\vskip 0.05in
\centering
\resizebox{\textwidth}{!}{
\begin{tikzpicture}
    \node[rectangle, clip, rounded corners=2pt, label={[label distance=-0.2cm]-90:\large \textbf{\textsf{(A)}}}] (img1) at (0,0) {\includegraphics[width=0.18\textwidth]{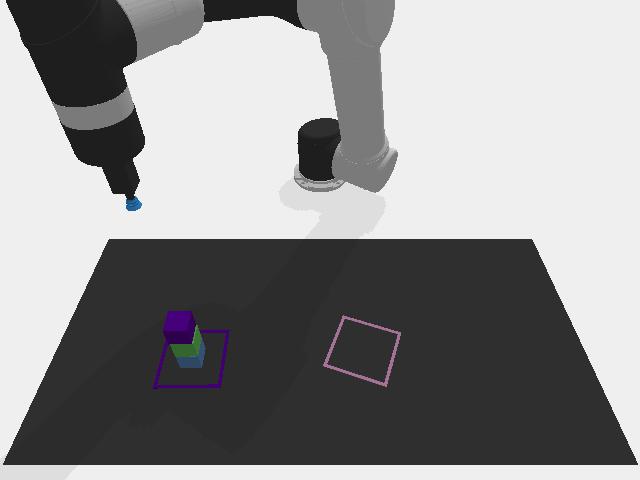}};
    
    \node[rectangle, rounded corners=2pt, clip, right=0.005cm of img1, label={[label distance=-0.2cm]-90:\large \textbf{\textsf{(B)}}}] (img2) {\includegraphics[width=0.18\textwidth]{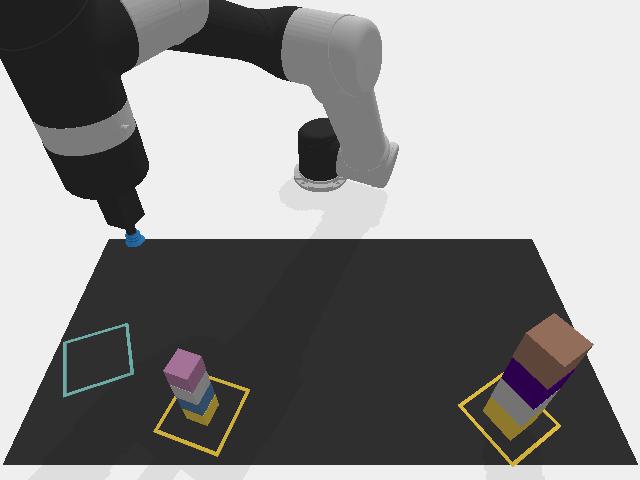}};
    
    \node[rectangle, rounded corners=2pt, clip, right=0.005cm of img2, label={[label distance=-0.2cm]-90:\large \textbf{\textsf{(C)}}}] (img3) {\includegraphics[width=0.18\textwidth]{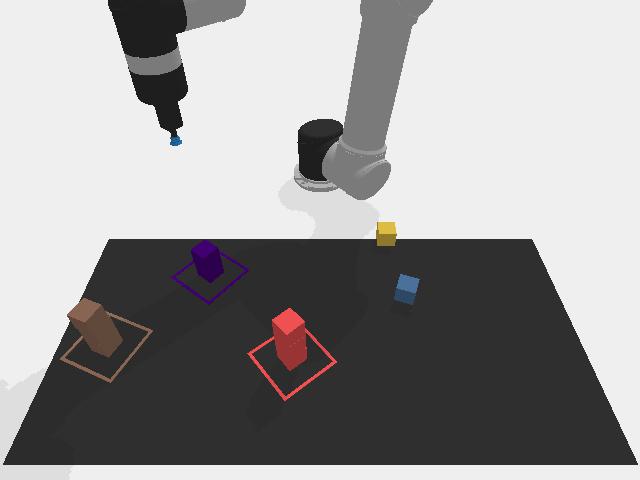}};
    
    \node[rectangle, rounded corners=2pt, clip, right=0.005cm of img3, , label={[label distance=-0.2cm]-90:\large \textbf{\textsf{(D)}}}] (img4) {\includegraphics[width=0.18\textwidth]{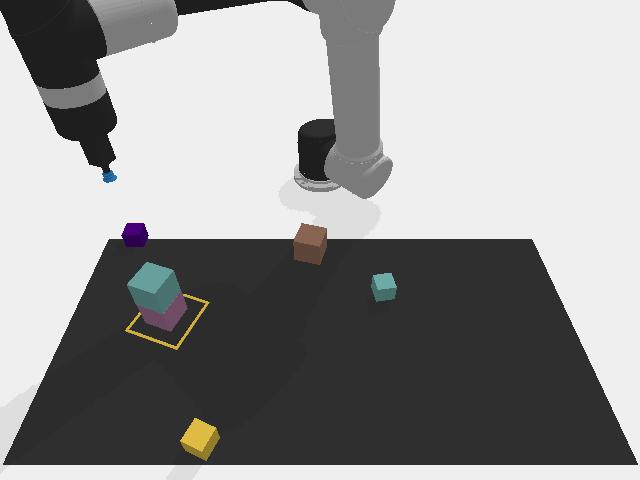}};
    
    \node[rectangle, rounded corners=2pt, clip, right=0.005cm of img4, label={[label distance=-0.2cm]-90:\large \textbf{\textsf{(E)}}}] (img5) {\includegraphics[width=0.18\textwidth]{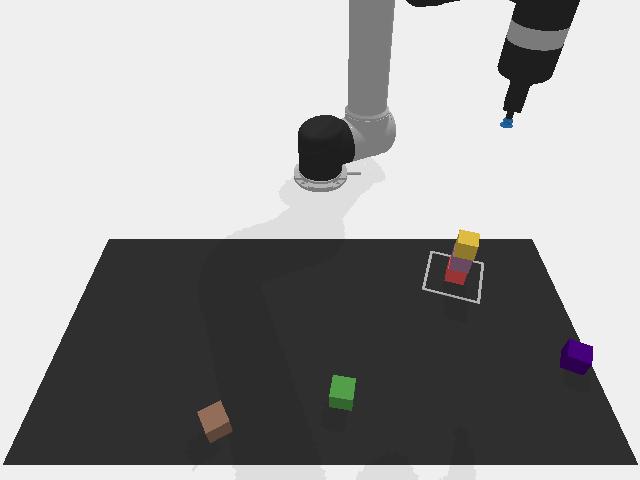}};
    
    \node[rectangle, rounded corners=2pt, clip, right=0.005cm of img5, label={[label distance=-0.2cm]-90:\large \textbf{\textsf{(F)}}}] (img6) {\includegraphics[width=0.18\textwidth]{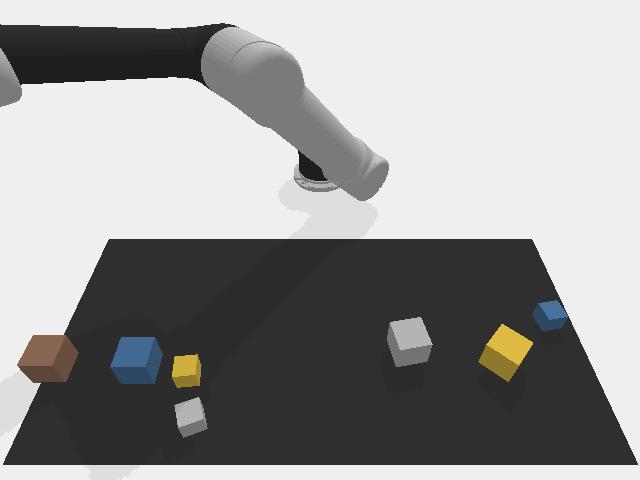}};
    
    \node[rectangle, rounded corners=2pt, clip, right=0.005cm of img6, label={[label distance=-0.2cm]-90:\large \textbf{\textsf{(G)}}}] (img7) {\includegraphics[width=0.18\textwidth]{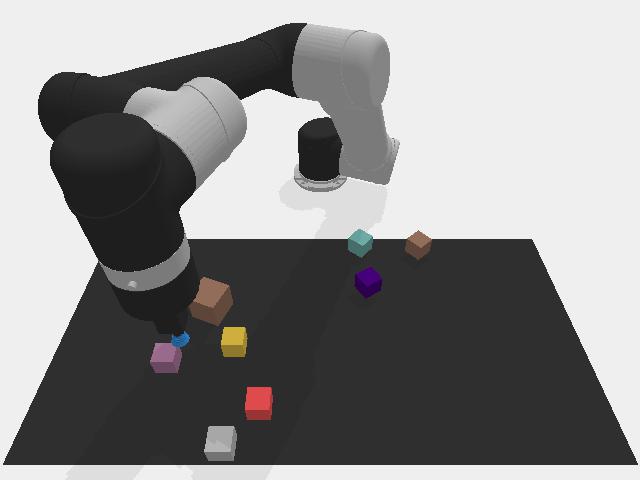}};
    
    \node[rectangle, rounded corners=2pt, clip, right=0.005cm of img7, label={[label distance=-0.2cm]-90:\large \textbf{\textsf{(H)}}}] (img8) {\includegraphics[width=0.18\textwidth]{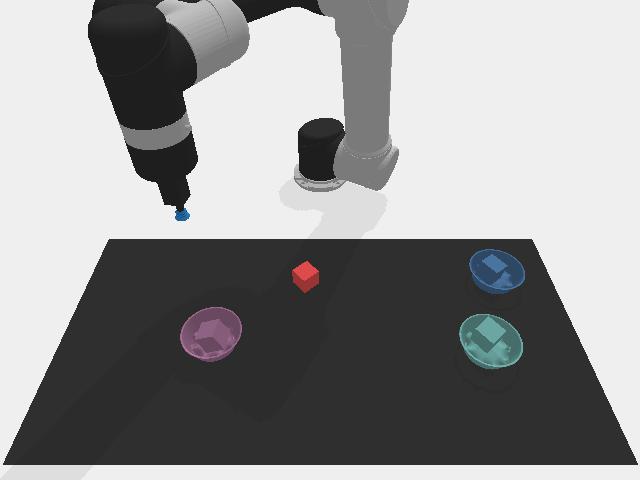}};
    
    \node[rectangle, rounded corners=2pt, clip, right=0.005cm of img8, label={[label distance=-0.2cm]-90:\large \textbf{\textsf{(I)}}}] (img9) {\includegraphics[width=0.18\textwidth]{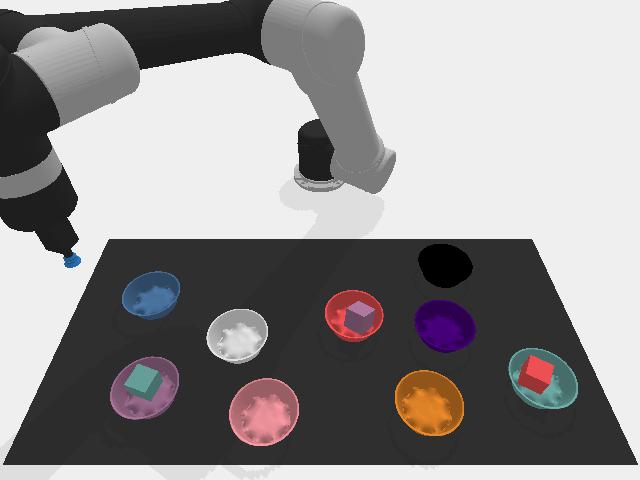}};
    
    \node[rectangle, rounded corners=2pt, clip, right=0.005cm of img9, label={[label distance=-0.2cm]-90:\large \textbf{\textsf{(J)}}}] (img10) {\includegraphics[width=0.18\textwidth]{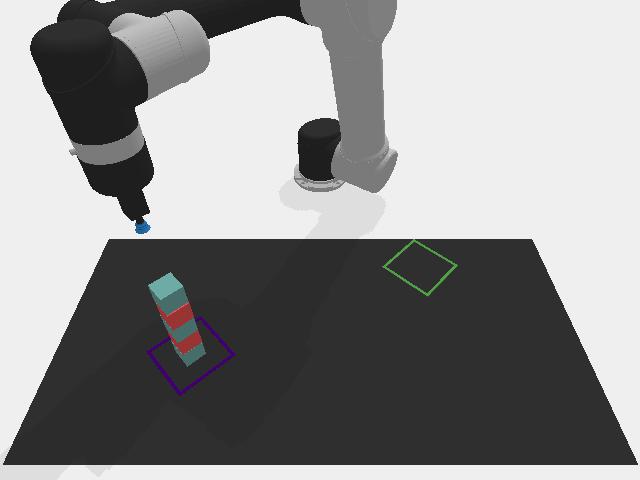}};
    
    \node[rectangle, rounded corners=2pt, clip, below=0.5cm of img1, label={[label distance=-0.2cm]-90:\large \textbf{\textsf{(G1)}}}] (img11) {\includegraphics[width=0.18\textwidth]{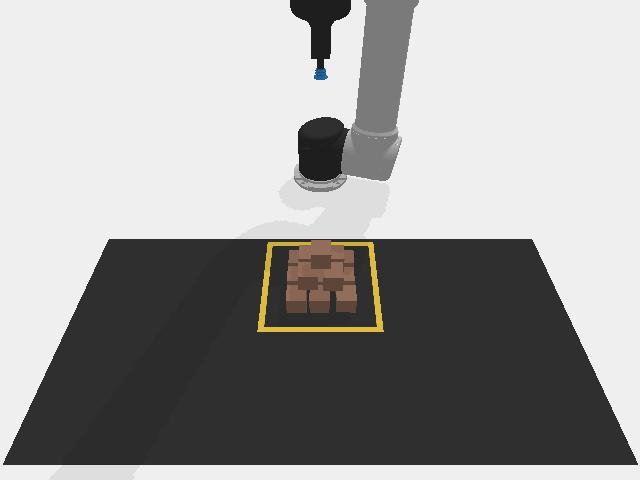}};
    
    \node[rectangle, rounded corners=2pt, clip, right=0.005cm of img11, label={[label distance=-0.2cm]-90:\large \textbf{\textsf{(G2)}}}] (img12) {\includegraphics[width=0.18\textwidth]{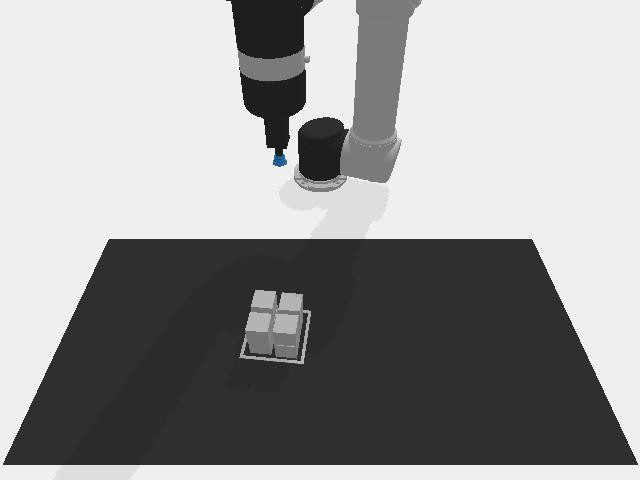}};
    
    \node[rectangle, rounded corners=2pt, clip, right=0.005cm of img12, label={[label distance=-0.2cm]-90:\large \textbf{\textsf{(G3)}}}] (img13) {\includegraphics[width=0.18\textwidth]{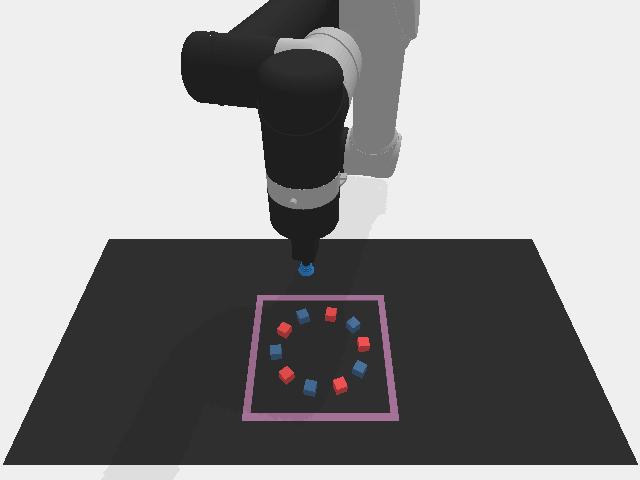}};
    
    \node[rectangle, rounded corners=2pt, clip, right=0.005cm of img13, label={[label distance=-0.2cm]-90:\large \textbf{\textsf{(G4)}}}] (img14) {\includegraphics[width=0.18\textwidth]{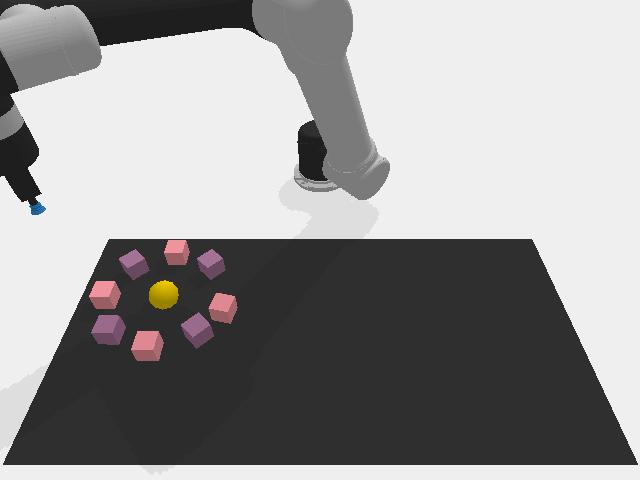}};
    
    \node[rectangle, rounded corners=2pt, clip, right=0.005cm of img14, label={[label distance=-0.2cm]-90:\large \textbf{\textsf{(G5)}}}] (img15) {\includegraphics[width=0.18\textwidth]{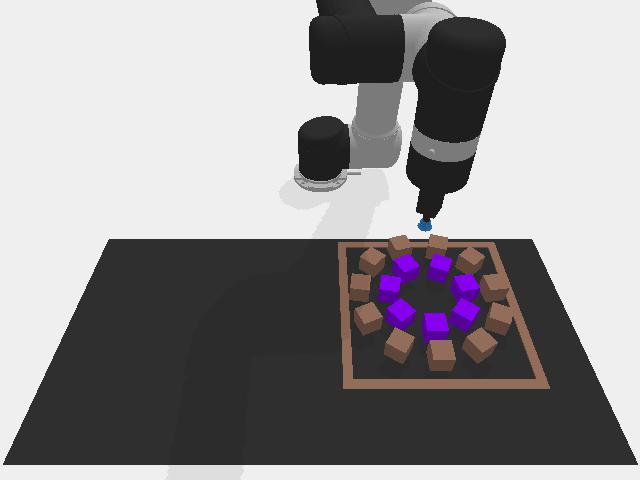}};
    
    \node[rectangle, rounded corners=2pt, clip, right=0.005cm of img15, label={[label distance=-0.2cm]-90:\large \textbf{\textsf{(G6)}}}] (img16) {\includegraphics[width=0.18\textwidth]{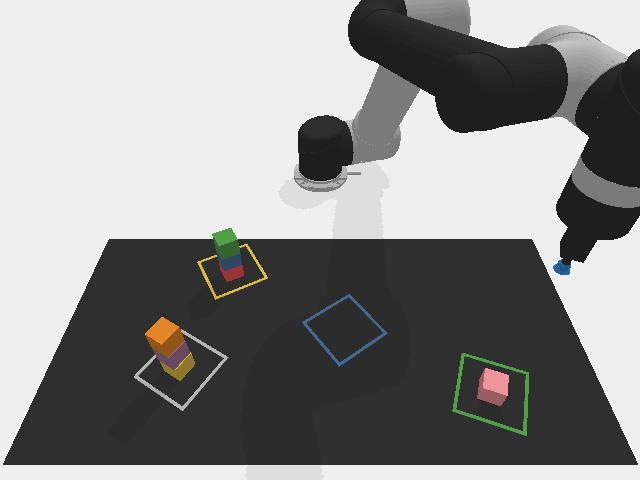}};
    
    \node[rectangle, rounded corners=2pt, clip, right=0.005cm of img16, label={[label distance=-0.2cm]-90:\large \textbf{\textsf{(G7)}}}] (img17) {\includegraphics[width=0.18\textwidth]{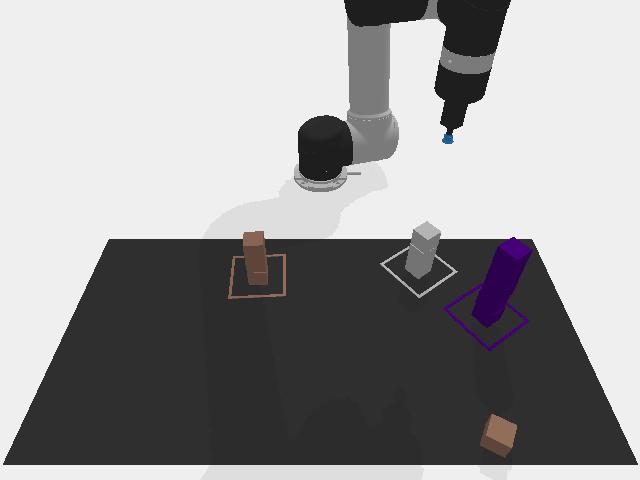}};
    
    \node[rectangle, rounded corners=2pt, clip, right=0.005cm of img17, label={[label distance=-0.2cm]-90:\large \textbf{\textsf{(G8)}}}] (img18) {\includegraphics[width=0.18\textwidth]{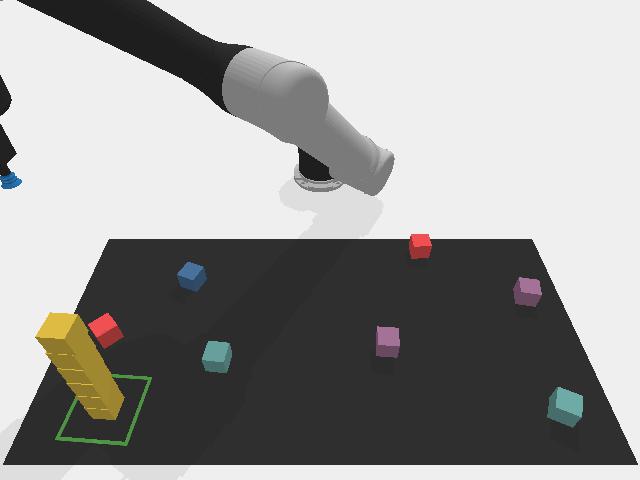}};
    
    \node[rectangle, rounded corners=2pt, clip, right=0.005cm of img18, label={[label distance=-0.2cm]-90:\large \textbf{\textsf{(G9)}}}] (img19) {\includegraphics[width=0.18\textwidth]{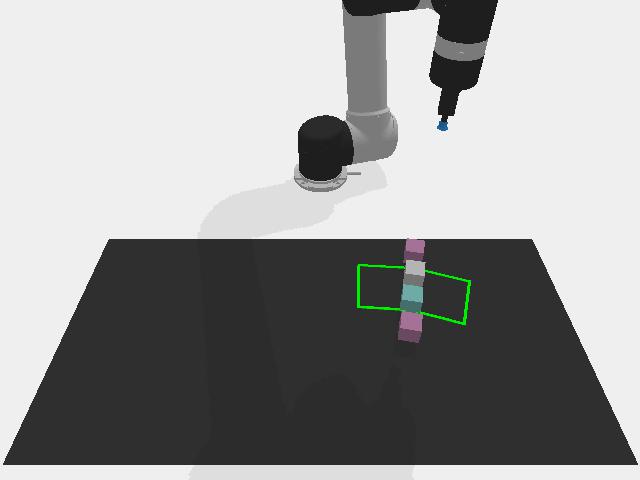}};
    
    \node[rectangle, rounded corners=2pt, clip, right=0.005cm of img19, label={[label distance=-0.2cm]-90:\large \textbf{\textsf{(G10)}}}] (img20) {\includegraphics[width=0.18\textwidth]{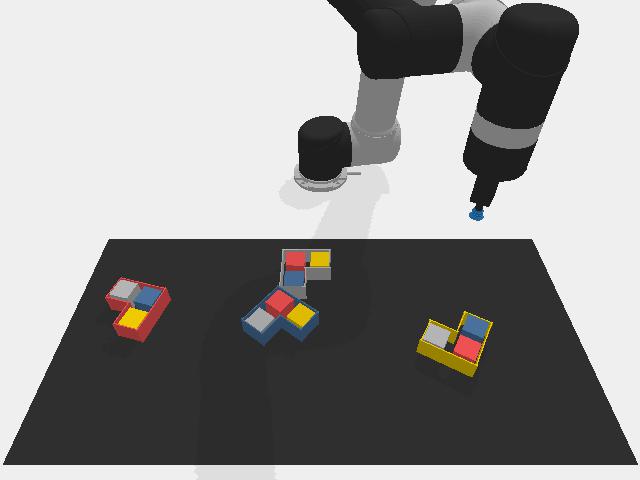}};

    \node[below right] at([yshift=-0.5cm]img11.south west) (table) {
    \begin{tabular}{c m{15.4cm} | c m{15.4cm}}
        \toprule
        \rowcolor{nature_tab_gray1}
        \textbf{Label} & \textbf{Instructions of Original Tasks} & \textbf{Label} & \textbf{Instructions of Generated Tasks} \\
        \midrule
        A & "Stack all blocks in the [COLOR] zone." & G1 & "Construct a 9-4-1 rectangular pyramid structure in the zone using 14 blocks of the same color." \\
        \rowcolor{nature_tab_gray2}
        \midrule
        B & "Stack blocks of the same size in the [COLOR1] zone and [COLOR2] zone respectively." & G2 & "Construct a 2*2*2 cube structure in the zone using 8 blocks of the same color." \\
        \midrule
        C & "Stack all the blocks of the same color together in the same colored zone." & G3 & "Construct a circle with suitable radius with alternating [COLOR1] and [COLOR2] blocks in the zone." \\
        \midrule
        \rowcolor{nature_tab_gray2}
        D & "Stack only the [SIZE] blocks of [COLOR\_TYPE] color in the [COLOR] zone." & G4 & "Construct a circle with suitable radius with alternating [COLOR1] and [COLOR2] blocks around the ball." \\
        \midrule
        E & "Stack all the blocks, which are to the [REL\_POS] of the [COLOR1] block with [POS\_TYPE] distance larger than 0.05 unit, in the [COLOR2] zone." & G5 & "Construct two concentric circles in the zone using [NUM] [COLOR1] and [NUM + 4] [COLOR2] blocks." \\
        \midrule
        \rowcolor{nature_tab_gray2}
        F & "Move all the blocks in the [POS1] area to [POS2] area." & G6 & "Divide the blocks into groups of [NUM] and stack each group (also including the group with block number less than [NUM]) in a different zone." \\
        \midrule
        G & "Move all the [SIZE] blocks in the [POS1] area to [POS2] area." & G7 & "Place the maximal odd number of blocks of the same color in each correspondingly colored zone." \\
        \rowcolor{nature_tab_gray2}
        \midrule
        H & "Put the blocks in the bowls with matching colors." & G8 & "Stack blocks of the same color that has the largest quantity in the zone." \\
        \midrule
        I & "Put the blocks in the bowls with mismatching colors." & G9 & "Arrange all blocks on the zone bisector line between two symmetrically placed zones evenly on the tabletop, and the gap between two adjacent blocks' edges should be near the block size, and the line connecting the center of the zones also bisects these blocks." \\
        \midrule
        \rowcolor{nature_tab_gray2}
        J & "Stack blocks with alternate colors on the [COLOR1] zone, starting with the [COLOR2] color." & G10 & "Each L-shaped fixture can hold three blocks, suppose the block size is (a,a,a), then in fixture's local coordinate system, the three places that can hold blocks are [(0,0,0),(a,0,0),(0,a,0)]. Fill in all the fixtures which have random position and rotation with blocks, and make sure in the end in every fixture there are three blocks with different colors." \\
        \bottomrule
    \end{tabular}
};
\end{tikzpicture}
}
\vskip -0.05in
\caption{Completion snapshots of LoHoRavens tasks. A-J are original tasks from LoHoRavens, and G1-G10 demonstrate generated tasks. The table shows the corresponding task instructions.}
\label{fig:snapshots}
\vskip -0.25in
\end{figure*}

\section{EXPERIMENT}

\textbf{Experiment setup:}
To evaluate our framework's long-horizon manipulation performance and scalability across diverse scenarios, we utilize 22 long-horizon tasks from three benchmarks: LoHoRavens \cite{zhang2023lohoravens}, CALVIN \cite{mees2022calvin}, and Franka Kitchen \cite{gupta2020relay}.
LoHoRavens, built on the Ravens, focuses on tabletop manipulation using a UR5e robot with a suction gripper. 
The environment features objects like blocks, bowls, and zones with variations in color, size, and texture. 
As shown in Fig. \ref{fig:snapshots}, we expand the LoHoRavens task pool to 20 tasks using GenSim \cite{wang2023gensim}, including 10 tasks from the original pool (A-J) and 10 newly generated, more challenging out-of-distribution tasks (G1-G10) for generalization evaluation, which are unseen in prompts or the training set. 
CALVIN offers four indoor setups featuring a Franka robot interacting with objects like drawers, sliding doors, buttons, and colored blocks. 
Similarly, the Franka Kitchen provides a kitchen scene where a Franka operates microwaves, burners, lights, and cabinets, etc. 
For both CALVIN and Franka Kitchen, we modify their base code setup (e.g., action space) and provide the necessary interfaces.
For real-world experiments, we evaluate our framework on a Franka robot with a RealSense D456 camera mounted at the table edge. 
To acquire the object identities and positions, we use Grounded SAM2 \cite{ren2024grounded}.

We adopt OpenAI’s GPT-4o-mini \cite{achiam2023gpt} as both the planner and reporter in our framework. 
For comparison in the code-generation domain, we evaluate against the representative open-source pipeline, Code as Policy (CaP).
By introducing randomly ordered in-context examples from the original task pool (Fig. \ref{fig:snapshots} A-J) into CaP, we assess the impact of our structured few-shot examples arranged in progressively increasing difficulty.
In the language-generation domain, we compare against state-of-the-art baselines from LoHoRavens, including the oracle CLIPort \cite{shridhar2022cliport}. 
Key distinctions between these baselines and our method include:
(1) Baselines perform task decomposition using natural language and rely on pretrained CLIPort \cite{shridhar2022cliport} as an action expert whose action space is end-effector Cartesian space.
In contrast, our framework does not depend on pretrained action policies; instead, it treats executable code generated by the LLM as the action space.
(2) Baselines rely on per-step reasoning, making them more prone to noise and cumulative errors, and less effective in long-horizon tasks. 
In contrast, our method executes all primitives in an episodic loop, and feedbacks are provided at inter-loop, improving both efficiency and robustness. 
(3) The baselines use randomly selected in-context examples from the task pool without structured reasoning guidance. 
Our method, however, organizes examples with progressively increasing complexity and employs CoT prompting to help the LLM extract underlying task logic and common knowledge, enabling better generalization and more reliable planning.
These baseline variations are thoroughly compared with our framework in the given tasks shown in Fig. \ref{fig:snapshots}.
Additionally, we conduct ablation studies to examine the individual contributions of task-level episodic feedback and progressively structured in-context examples with CoT reasoning. Specifically, we compare the full DAHLIA framework with (i) a planner-only variant that operates in an open-loop manner (DAHLIA (planner-only)) but with the same in-context examples setup as the original framework, and (ii) a variant that replaces progressive example ordering with randomly selected in-context examples and omits CoT guidance (DAHLIA (random)).
All tasks are tested with 50 random seeds, and average success rates (SR, \%) are reported.
Results show our closed-loop setup achieves the highest success rates across both seen and unseen tasks.



\textbf{Long-horizon manipulation:}
\begin{table}[b!]
\vskip -0.15in
    \centering   
    \caption{SR (\%) for tasks from the original LoHoRavens task pool.}
    \vskip -0.05in
    \resizebox{.49\textwidth}{!}{
    \begin{threeparttable}
    \begin{tabular}{l cccccccccc}
        \toprule
        \rowcolor{nature_tab_gray1}
         & \multicolumn{10}{c}{\textbf{Long-horizon Task Average Success Rate (\%)}}\\
        \rowcolor{nature_tab_gray1}
        \textbf{Frameworks} & A & B & C & D & E & F & G & H & I & J \\
        \midrule
        CLIPort (oracle) & 22 & 2 & 8 & 2 & 2 & 16 & 10 & 20 & 18 & 2 \\
        \rowcolor{nature_tab_gray2}
        \midrule
        $^*$Llama2$^P$ + Flamingo$^R$ $\rightarrow$ CLIPort & 20 & 10 & 4 & 10 & 16 & 28 & 32 & 32 & 28 & 14 \\
        \midrule
        $^*$Llama3$^P$ + CogVLM2$^R$ $\rightarrow$ CLIPort & 20 & 20 & 24 & 22 & 14 & 20 & 20 & 32 & 30 & 22 \\
        \rowcolor{nature_tab_gray2}
        \midrule
        $^*$GPT-4o-mini$^P$ + CogVLM2$^R$ $\rightarrow$ CLIPort & 26 & 22 & 30 & 24 & 20 & 34 & 28 & 36 & 36 & 30 \\
        \midrule
        $^*$GPT-4o-mini$^P$ + GPT-4o-mini$^R$ $\rightarrow$ CLIPort & 30 & 32 & 32 & 30 & 18 & 34 & 30 & 46 & 36 & 32 \\
        \rowcolor{nature_tab_gray2}
        \midrule
        CaP (GPT-4o-mini) & 86 & 32 & 42 & 36 & 10 & 60 & 38 & 30 & 22 & 26\\
        \midrule
        DAHLIA (planner-only) & 100 & 88 & 96 & 60 & 42 & 76 & 80 & 90 & 90 & 82\\
        \rowcolor{nature_tab_gray2}
        \midrule
        DAHLIA (random) & 98 & 96 & 98 & 50 & 46 & 68 & 80 & 92 & 94 & 80\\
        \midrule
        \textbf{DAHLIA (Ours)} & 100 & 100 & 100 & 48 & 42 & 88 & 72 & 100 & 90 & 80\\
    \bottomrule
    \end{tabular}
    \begin{tablenotes}[flushleft]
        \normalsize
        \item[]
        $^*$Adopted from the LoHoRavens \cite{zhang2023lohoravens}, where all baselines use planner for per-step inference, reporter for feedback, and CLIPort for action execution.
        $^P$Planner model. $^R$Reporter model.
    \end{tablenotes}
    \end{threeparttable}
    }
    \label{tab:success}
\end{table}
We first evaluate all models on the original LoHoRavens tasks (Fig. \ref{fig:snapshots} A-J). 
As shown in Table \ref{tab:success}, our proposed framework, DAHLIA, outperforms all baselines in long-horizon manipulation tasks, achieving the highest SR, including 100\% accuracy in five out of ten tasks. 
Compared to its planner-only setup and open-loop CaP, our dual-tunnel setup shows clear advantages, especially in tasks requiring precise coordination and iterative re-planning.
While CaP performs well on simple tasks like stacking blocks (86\% SR), its performance drops substantially on more challenging long-horizon tasks due to its use of randomly ordered in-context examples, which limits the LLM’s ability to generalize and reason over spatial and contextual relationships.
To isolate the effect of structured in-context learning, we further compare DAHLIA against DAHLIA (random).
Although DAHLIA (random) consistently outperforms CaP and the planner-only setup, it underperforms the original DAHLIA framework on several long-horizon tasks, particularly those involving complex spatial constraints and multi-step dependencies (e.g., Tasks D, F, and G).
This performance gap highlights the importance of progressively structured few-shot examples and CoT reasoning, which enable the planner to internalize reusable task abstractions and maintain coherent plan structure over long horizons.
Other language generation baselines using per-step inference with CLIPort also struggle. 
For instance, GPT-4o-mini$^P$ + GPT-4o-mini$^R$ $\rightarrow$ CLIPort achieves relatively higher SR than others but suffers from cumulative errors, resulting in an average SR below 32\%. 
In contrast, DAHLIA leverages CoT-guided in-context learning with progressively arranged examples and integrates closed-loop feedback, leading to up to 12\% performance gains and effectively reducing error accumulation and ensuring robust long-horizon task execution.

\begin{table}[b!]
\vskip -.1in
    \centering
    \caption{SR (\%) for unseen out-of-distribution long-horizon tasks.}
    \vskip -0.05in
    \resizebox{.49\textwidth}{!}{
    \begin{threeparttable}
    \begin{tabular}{l cccccccccc}
        \toprule
        \rowcolor{nature_tab_gray1}
         & \multicolumn{10}{c}{\textbf{Long-horizon Task Average Success Rate (\%)}}\\
        \rowcolor{nature_tab_gray1}
        \textbf{Frameworks} & G1 & G2 & G3 & G4 & G5 & G6 & G7 & G8 & G9 & G10\\
        \midrule
        CLIPort (oracle) & 0 & 32 & 4 & 0 & 0 & 0 & 0 & 8 & 0 & 0\\
        \midrule
        \rowcolor{nature_tab_gray2}
        $^*$GPT-4o-mini$^P$ + GPT-4o-mini$^R$ $\rightarrow$ CLIPort & 8 & 36 & 34 & 30 & 4 & 28 & 30 & 18 & 0 & 0 \\
        \midrule
        CaP (GPT-4o-mini) & 6 & 40 & 32 & 38 & 0 & 0 & 44 & 46 & 0 & 0 \\
        \midrule
        \rowcolor{nature_tab_gray2}
        DAHLIA (planner-only) & 58 & 76 & 92 & 100 & 42 & 98 & 88 & 88 & 12 & 8 \\
        \midrule
        DAHLIA (random) & 36 & 38 & 40 & 90 & 50 & 90 & 72 & 66 & 0 & 0 \\
        \midrule
        \rowcolor{nature_tab_gray2}
        \textbf{DAHLIA (Ours)} & 50 & 100 & 100 & 100 & 50 & 100 & 100 & 90 & 18 & 10 \\
    \bottomrule
    \end{tabular}
    \begin{tablenotes}[flushleft]
        \normalsize
        \item
        $^*$Adopted from the LoHoRavens \cite{zhang2023lohoravens}, where all baselines use planner for per-step inference, reporter for feedback, and CLIPort for action execution. $^P$Planner model. $^R$Reporter model. 
    \end{tablenotes}
    \end{threeparttable}
    }
    \label{tab:success_gen_task}
\end{table}

\textbf{Out-of-distribution performance:} To further assess the generalization performance in more challenging out-of-distribution scenarios, we evaluate models models on a newly generated set of long-horizon tasks that are unseen during prompting (Fig. \ref{fig:snapshots}, G1-G10). No task-specific examples from this set are included in the in-context prompts or dataset, ensuring that these scenarios are novel to all models.
The results in Table \ref{tab:success_gen_task} demonstrate that DAHLIA consistently achieves the highest SR, further validating its robust generalization in challenging unseen scenarios.
In contrast, baseline models experience a notable decline in SR when generalized to unseen scenarios.
Comparing DAHLIA with its planner-only variant highlights the importance of episodic closed-loop feedback.
Incorporating outcome evaluation and replanning leads to substantial improvements on several tasks, including G2 (+24\%), G3 (+8\%), G5 (+8\%), and G7 (+12\%), underscoring the role of iterative evaluation in recovering from execution errors.
More importantly, the comparison between DAHLIA and DAHLIA (random) isolates the effect of progressively structured few-shot examples and CoT-guided reasoning.
Although DAHLIA (random) performs competitively on some tasks that closely resemble those seen during prompting (e.g., G4 and G6), its performance degrades sharply on more diverse or compositional unseen tasks (e.g., G2, G3, G7, G9, and G10).
This degradation indicates that, without incremental example ordering and explicit reasoning guidance, the model tends to rely on superficial alignment with previously seen patterns and lacks a deeper understanding of task structure.

In the absence of progressive few-shot conditioning, DAHLIA (random) primarily learns to match examples at a surface level, which suffices for familiar tasks but fails to transfer to genuinely novel configurations.
When confronted with OOD scenarios that require recombining known skills in unfamiliar ways, the planner struggles to adapt due to insufficient internalization of reusable planning abstractions.
In contrast, progressively ordered examples combined with CoT reasoning encourage the model to focus on task decomposition and compositional logic, enabling effective generalization across unseen task distributions.

Finally, closed-loop episodic feedback further enhances robustness by allowing the system to detect and correct failures after execution, leading to consistently higher SR in DAHLIA.
Together, these results confirm that incremental few-shot learning and CoT-guided reasoning are critical for robust generalization in long-horizon code generation, particularly in out-of-distribution robotic long-horizon manipulation tasks.

%

\begin{figure}[t!]
\centering
\resizebox{.5\textwidth}{!}{
\begin{tikzpicture}
    \node[rectangle, clip, rounded corners=2pt, label={[label distance=-0.6cm]-90:\small \textbf{\textsf{lift block}}}] (img1) at (0,0) {\includegraphics[width=0.18\textwidth]{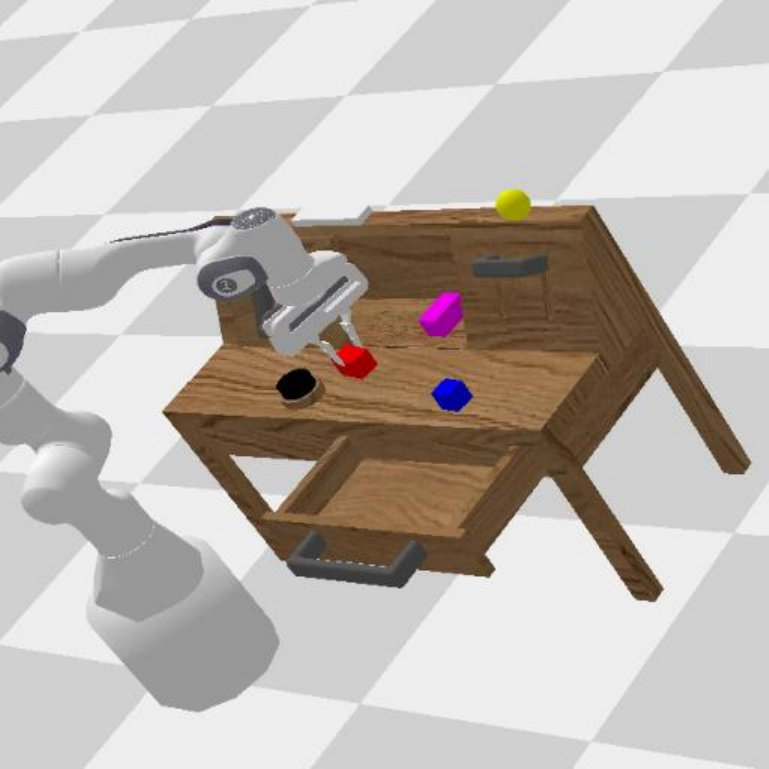}};

    \node[rectangle, rounded corners=2pt, clip, right=0.005cm of img1, label={[label distance=-0.6cm]-90:\small \textbf{\textsf{rotate block}}}] (img2) {\includegraphics[width=0.18\textwidth]{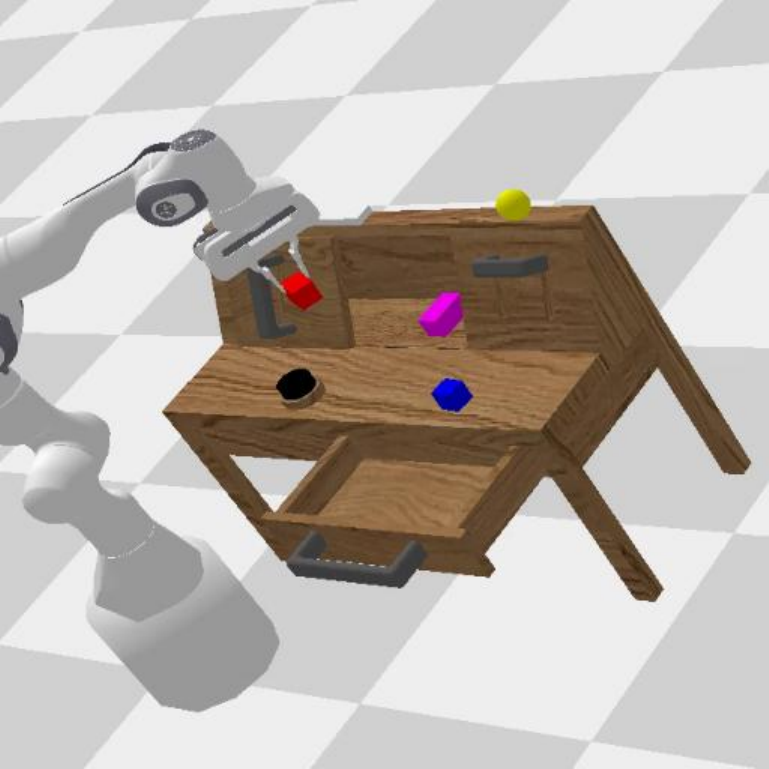}};

    \node[rectangle, rounded corners=2pt, clip, right=0.005cm of img2, label={[label distance=-0.6cm]-90:\small \textbf{\textsf{turn on switch}}}] (img3) {\includegraphics[width=0.18\textwidth]{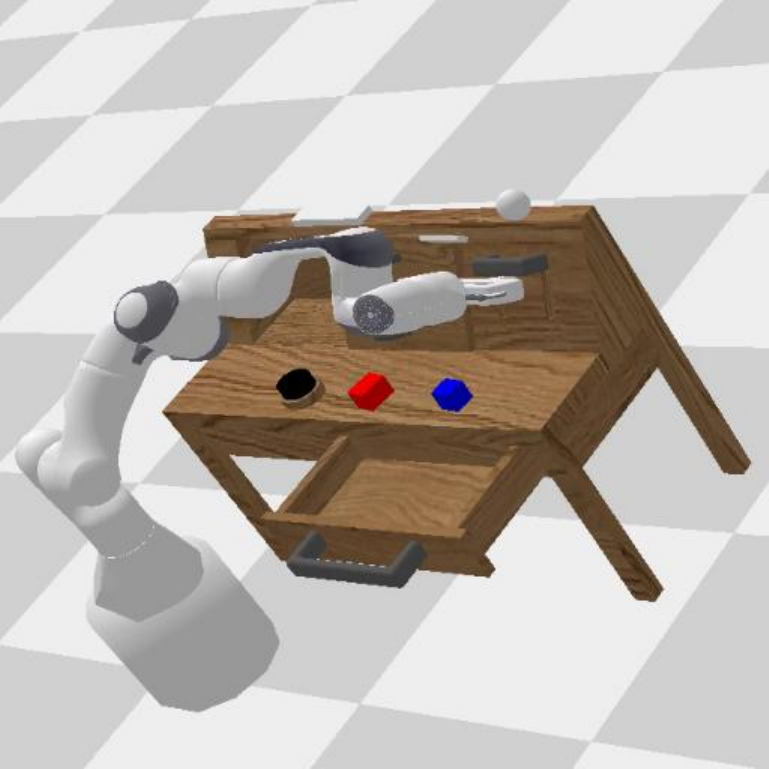}};

    \node[rectangle, rounded corners=2pt, clip, right=0.005cm of img3, label={[label distance=-0.6cm]-90:\small \textbf{\textsf{open slide door}}}] (img4) {\includegraphics[width=0.18\textwidth]{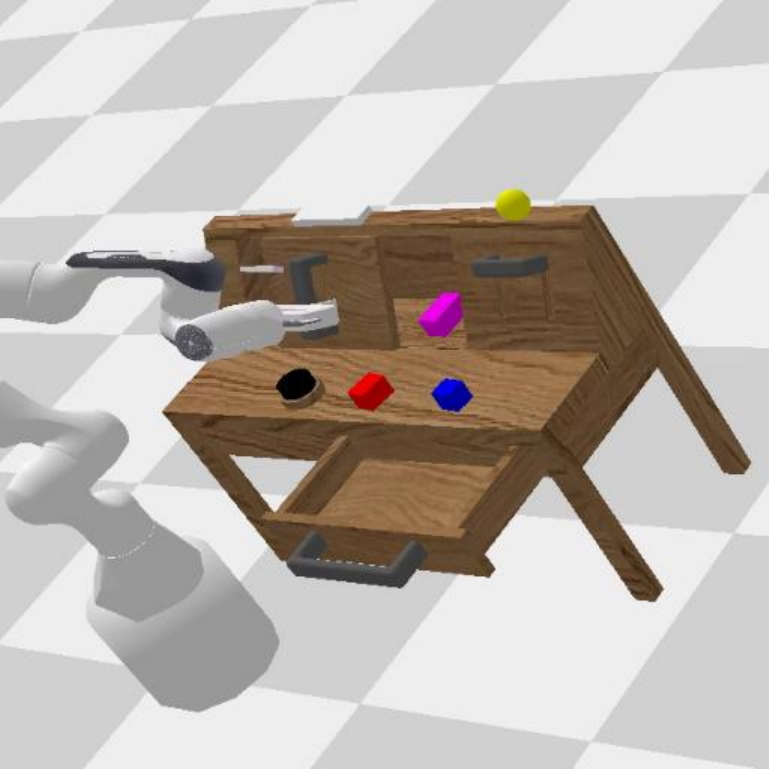}};

    \node[rectangle, rounded corners=2pt, clip, right=0.005cm of img4, label={[label distance=-0.6cm]-90:\small \textbf{\textsf{open drawer}}}] (img5) {\includegraphics[width=0.18\textwidth]{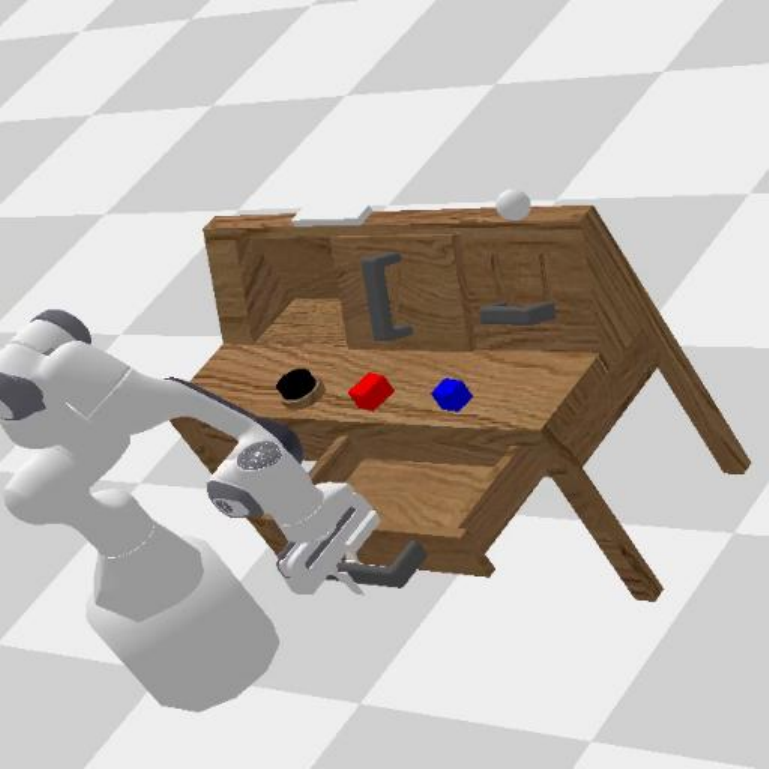}};

    \node[rectangle, clip, rounded corners=2pt, below=0.1cm of img1, label={[label distance=-0.6cm]-90:\small \textbf{\textsf{\textcolor{white}{microwave}}}}] (img6) {\includegraphics[width=0.18\textwidth]{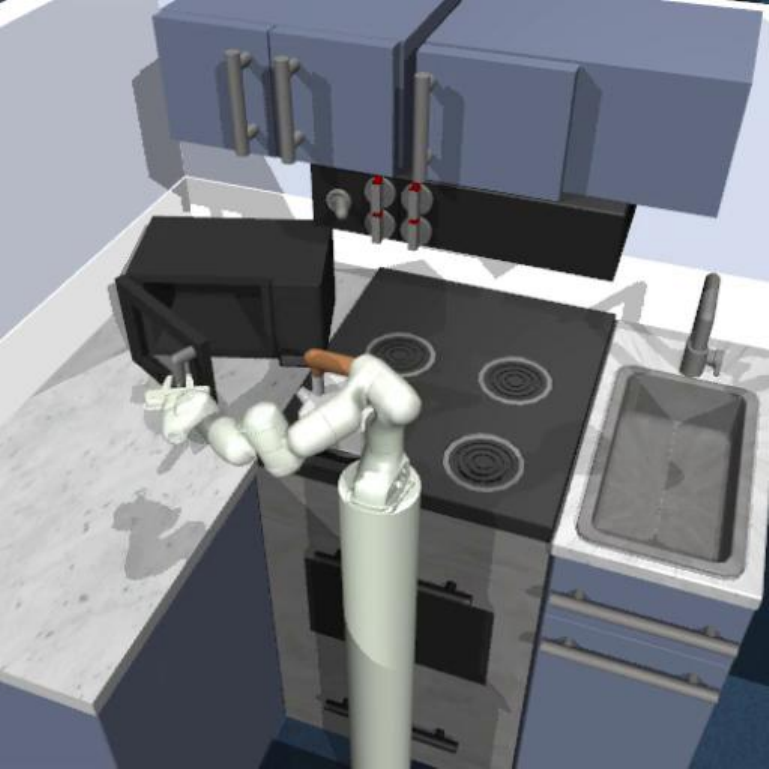}};

    \node[rectangle, rounded corners=2pt, clip, right=0.005cm of img6, label={[label distance=-0.6cm]-90:\small \textbf{\textsf{\textcolor{white}{kettle}}}}] (img7) {\includegraphics[width=0.18\textwidth]{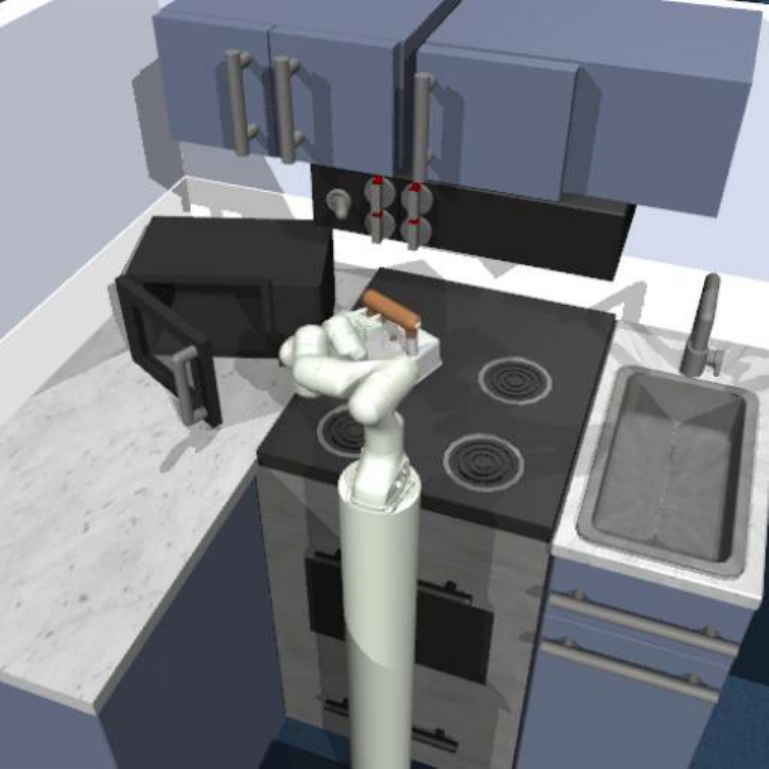}};

    \node[rectangle, rounded corners=2pt, clip, right=0.005cm of img7, label={[label distance=-0.6cm]-90:\small \textbf{\textsf{\textcolor{white}{burner}}}}] (img8) {\includegraphics[width=0.18\textwidth]{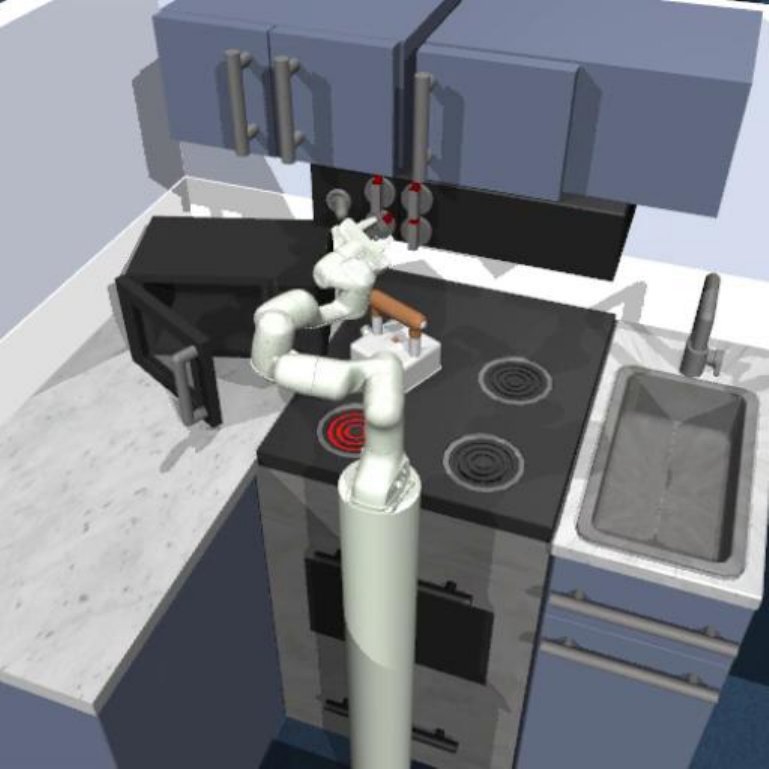}};

    \node[rectangle, rounded corners=2pt, clip, right=0.005cm of img8, label={[label distance=-0.6cm]-90:\small \textbf{\textsf{\textcolor{white}{slide cabinet}}}}] (img9) {\includegraphics[width=0.18\textwidth]{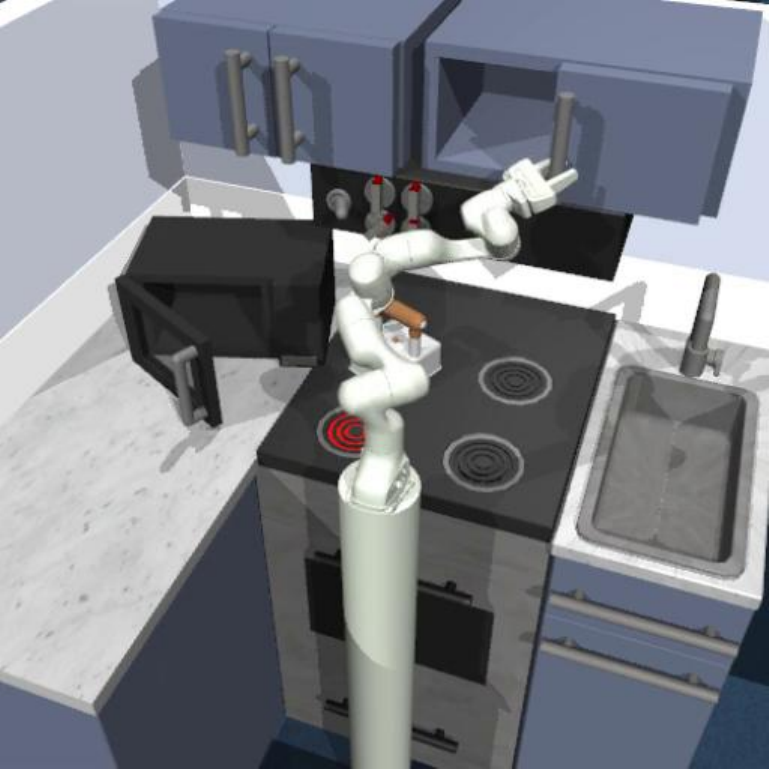}};

    \node[rectangle, rounded corners=2pt, clip, right=0.005cm of img9, label={[label distance=-0.6cm]-90:\small \textbf{\textsf{\textcolor{white}{hinge cabinet}}}}] (img10) {\includegraphics[width=0.18\textwidth]{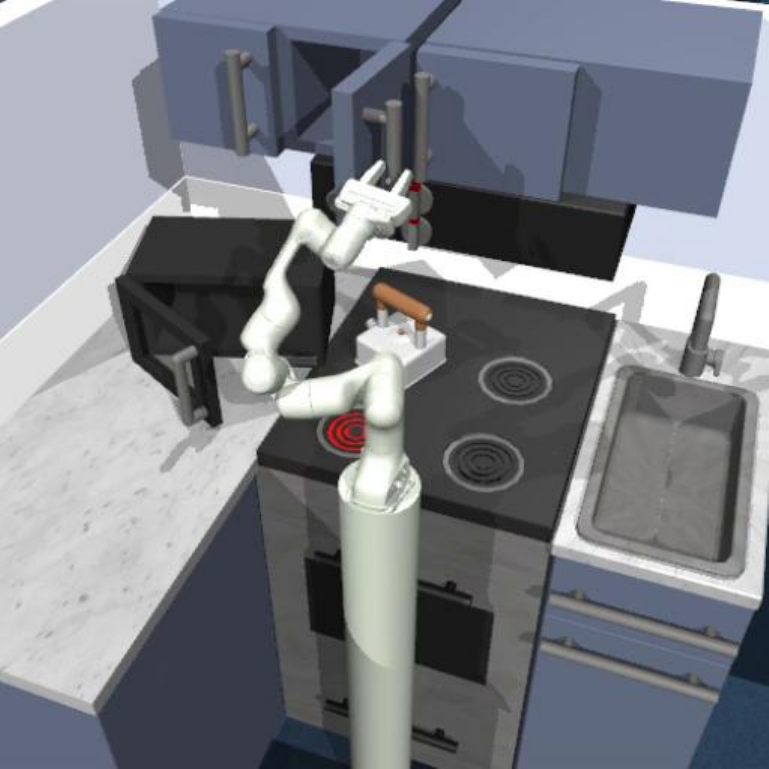}};

\end{tikzpicture}

}
\vskip -0.05in
\caption{Completion snapshots of CALVIN and Franka Kitchen benchmarks.}
\label{fig:calvin_kitchen_snapshots}
\vskip -0.2in
\end{figure}

\begin{table}[b!]
\vskip -.1in
    \centering
    \caption{SR (\%) for subtasks in CALVIN and Franka Kitchen}
    \vskip -0.05in
    \resizebox{.46\textwidth}{!}{
    \begin{threeparttable}
    \begin{tabular}{lcc}
    \toprule
    \rowcolor{nature_tab_gray1}
     \textbf{CALVIN Tasks} & DAHLIA (planner-only) & DAHLIA (Ours) \\
     \midrule
     lift block & 100 & 100 \\
     \rowcolor{nature_tab_gray2}
     \midrule
     rotate block & 94 & 98 \\
     \midrule
     turn on switch & 96 & 100 \\
     \rowcolor{nature_tab_gray2}
     \midrule
     open slide door & 100 & 100 \\
     \midrule
     open drawer & 100 & 100 \\
     \midrule
     \rowcolor{nature_tab_gray2}
     \textbf{overall} & 98.00 & 99.60 \\
     \midrule
     \rowcolor{nature_tab_gray1}
     \textbf{Franka Kitchen Tasks} & DAHLIA (planner-only) & DAHLIA (Ours) \\
     \midrule
     microwave & 98 & 100 \\
     \rowcolor{nature_tab_gray2}
     \midrule
     kettle & 100 & 100 \\
     \midrule
     light & 98 & 100 \\
     \rowcolor{nature_tab_gray2}
     \midrule
     top burner & 90 & 96 \\
     \midrule
     bottom burner & 90 & 98 \\
     \rowcolor{nature_tab_gray2}
     \midrule
     slider cabinet & 100 & 100 \\
     \midrule
     hinge cabinet & 96 & 98 \\
     \midrule
     \rowcolor{nature_tab_gray2}
     \textbf{overall} & 96.00 & 98.86 \\
    \bottomrule
    \end{tabular}
    \end{threeparttable}
    }
    \label{tab:scalability}
\end{table}
\textbf{Scalability:}
To assess the scalability and generalization of our framework across different embodiments and task setups, we conduct additional experiments in CALVIN and Franka Kitchen. 
We wrapped the underlying interface to make the naming and functionality consistent with previous APIs.
The average SR for each subtask is shown in Table \ref{tab:scalability}. The results show that our default dual-tunnel DAHLIA setup consistently outperforms its planner-only variant, demonstrating the effectiveness of its closed-loop feedback in improving execution reliability. 
Our framework achieves 100\% success in most tasks, including lift block, open drawer, open slide door, kettle, and slider cabinet, while also surpassing the planner-only model in complex tasks requiring precise coordination, such as rotate block (+4\%) and turn on switch (+4\%) in CALVIN, as well as top burner (+6\%) and bottom burner (+8\%) in Franka Kitchen.
Notably, the planner-only model may struggle in tasks requiring highly accurate dexterous manipulation. 
For example, in the ``top burner'' and ``bottom burner'' tasks in Franka Kitchen, the robot must accurately reach the correct rotary knob and rotate it by 45 degrees to open the burner. 
Errors in task planning or dexterous execution, such as selecting the wrong switch or rotating in the wrong direction, can lead to complete task failure. 
In such cases, refined structured feedback plays a critical role by detecting and correcting execution errors, enabling the planner to refine its actions (e.g., correct the contact pose or rotation direction).
The completion snapshots in Fig. \ref{fig:calvin_kitchen_snapshots} showcase the effectiveness of our framework across diverse long-horizon tasks, from structured tabletop interactions in CALVIN to articulated object handling in Franka Kitchen. These results further demonstrate our pipeline's ability to generalize across diverse task scenarios and embodiments. For more details, refer to our attached videos.
%

\begin{figure}[t!]
\centering
\resizebox{.5\textwidth}{!}{
\begin{tikzpicture}
    \node[rectangle, clip, rounded corners=2pt,] (img1) at (0,0) {\includegraphics[width=0.18\textwidth]{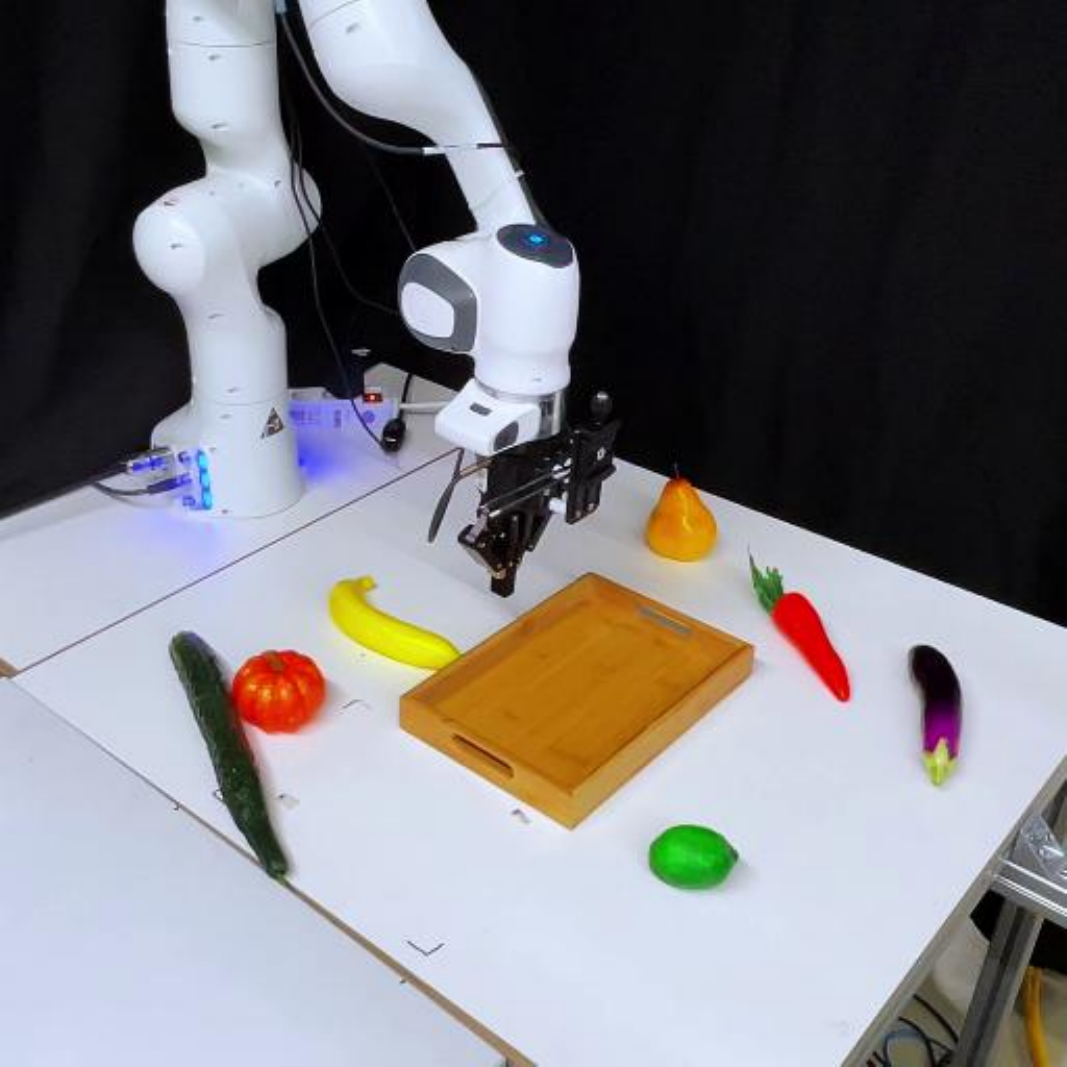}};

    \node[rectangle, rounded corners=2pt, clip, right=0.005cm of img1,] (img2) {\includegraphics[width=0.18\textwidth]{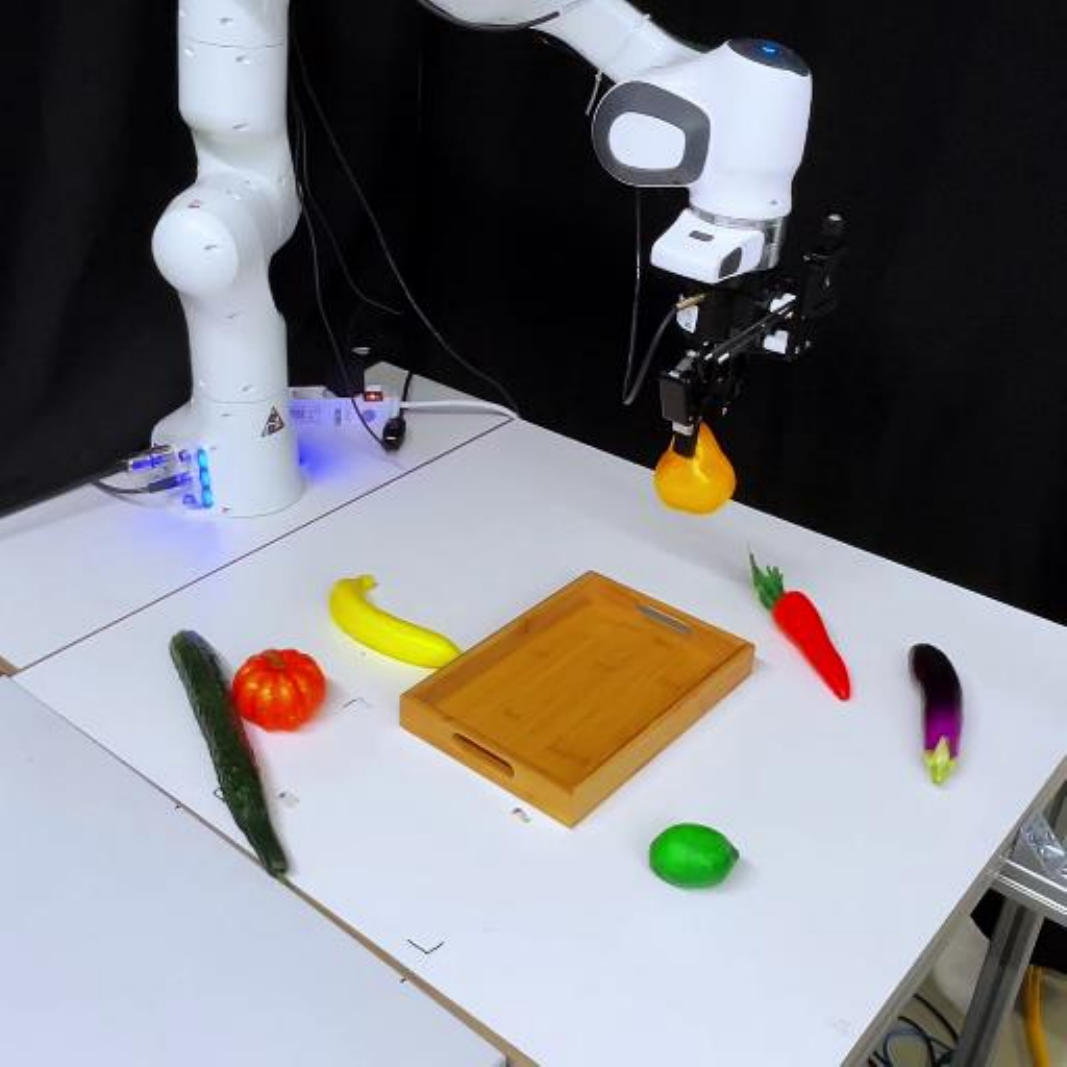}};

    \node[rectangle, rounded corners=2pt, clip, right=0.005cm of img2, ] (img3) {\includegraphics[width=0.18\textwidth]{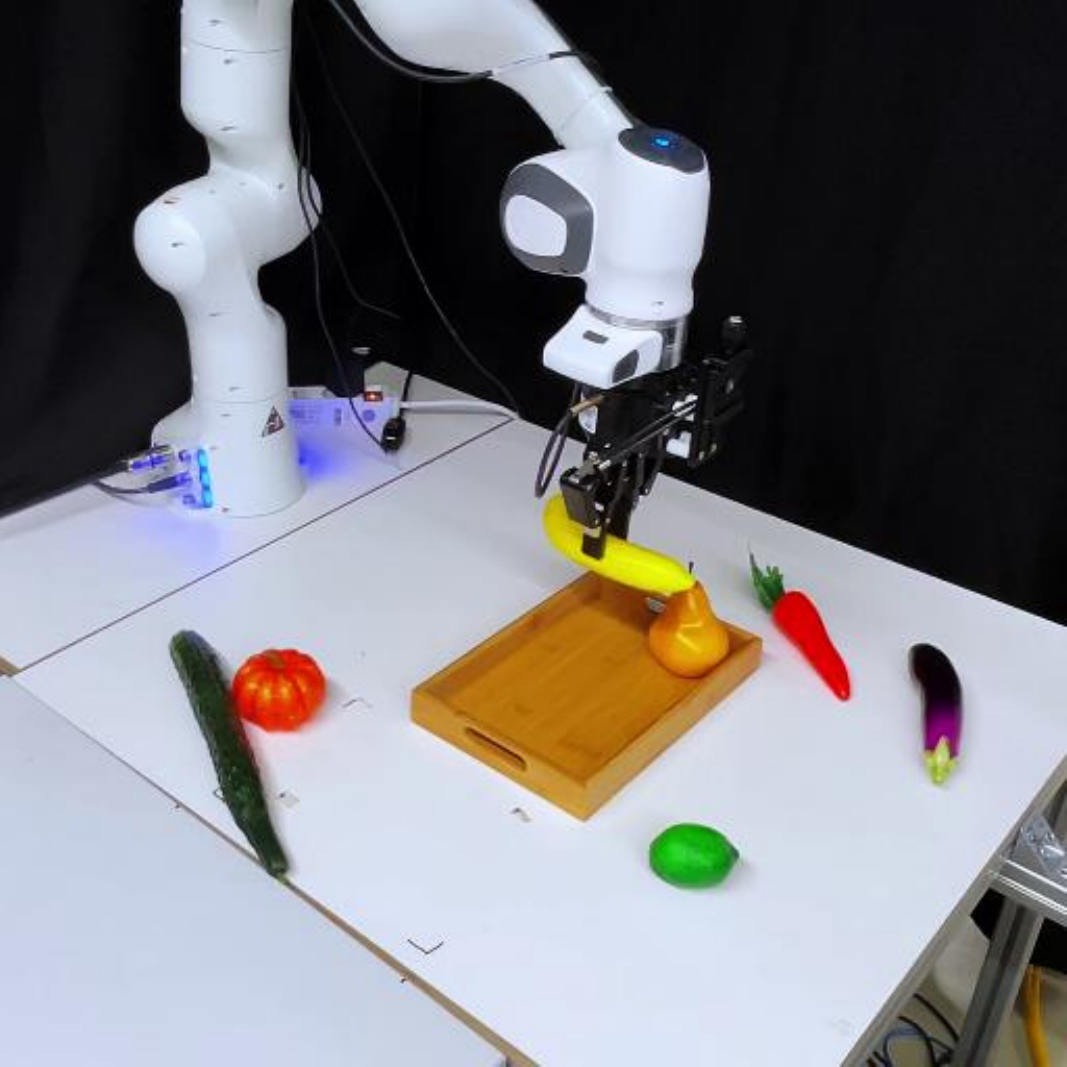}};

    \node[rectangle, rounded corners=2pt, clip, right=0.005cm of img3, ] (img4) {\includegraphics[width=0.18\textwidth]{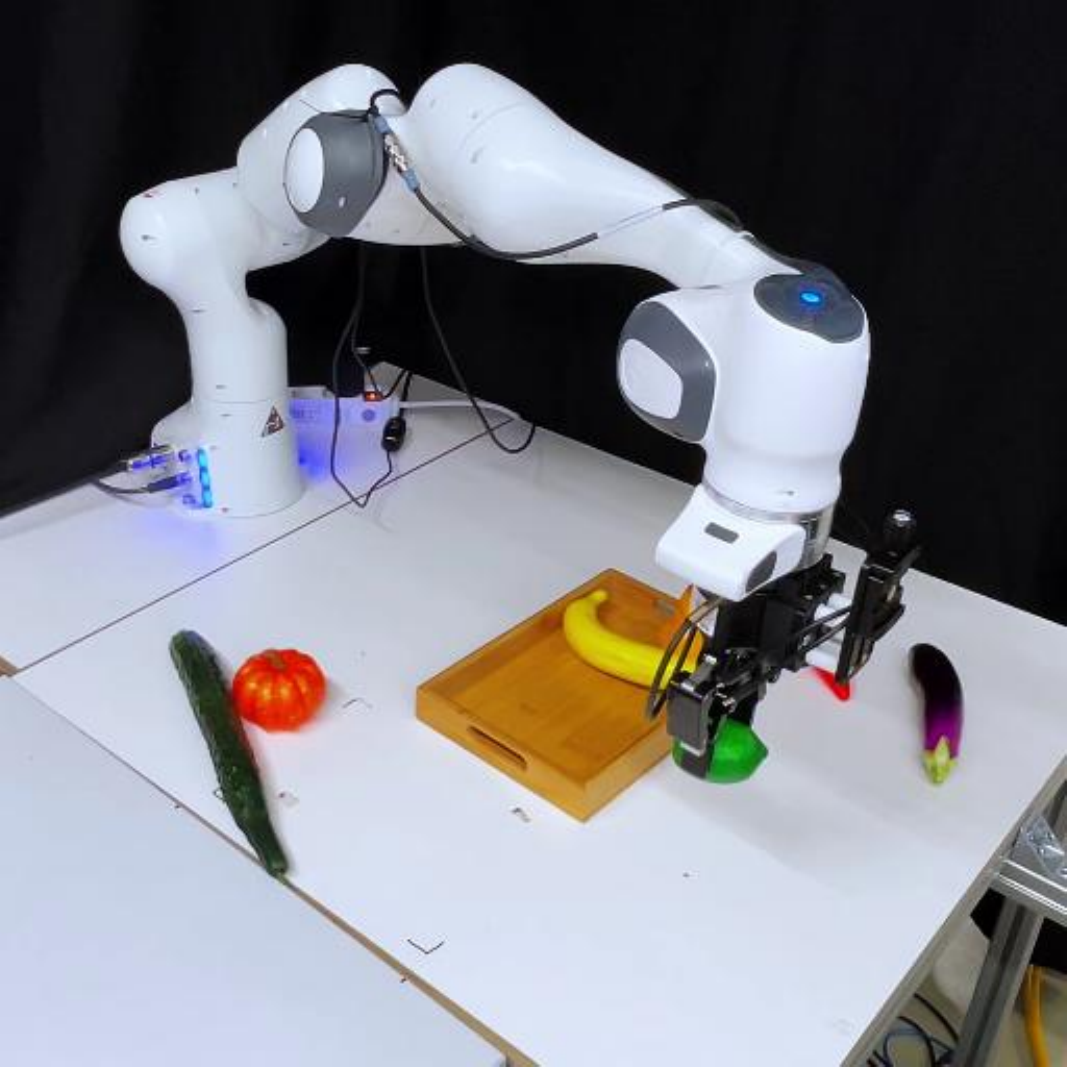}};

    \node[rectangle, rounded corners=2pt, clip, right=0.005cm of img4,] (img5) {\includegraphics[width=0.18\textwidth]{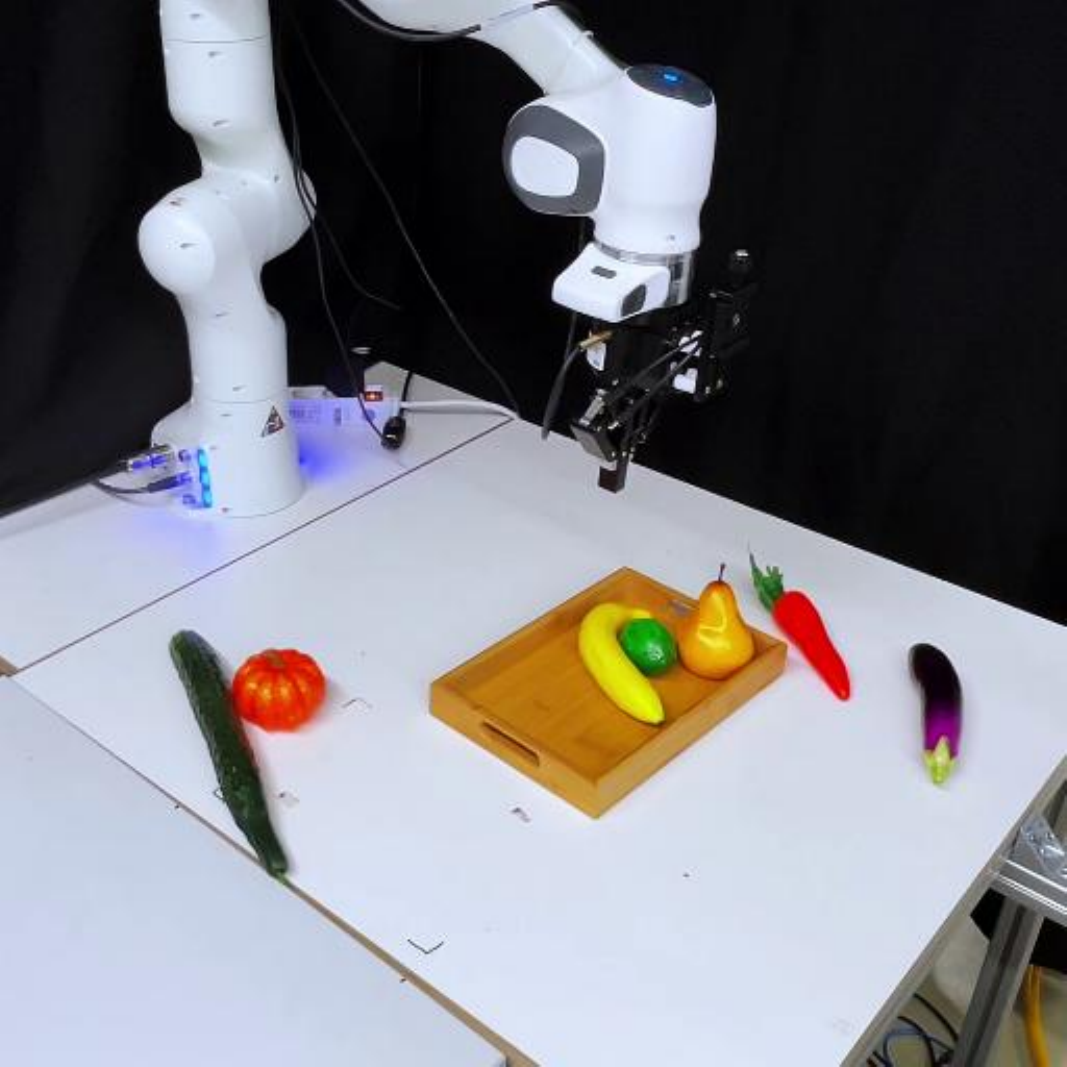}};

    \node[rectangle, clip, rounded corners=2pt, below=0.1cm of img1,] (img6) {\includegraphics[width=0.18\textwidth]{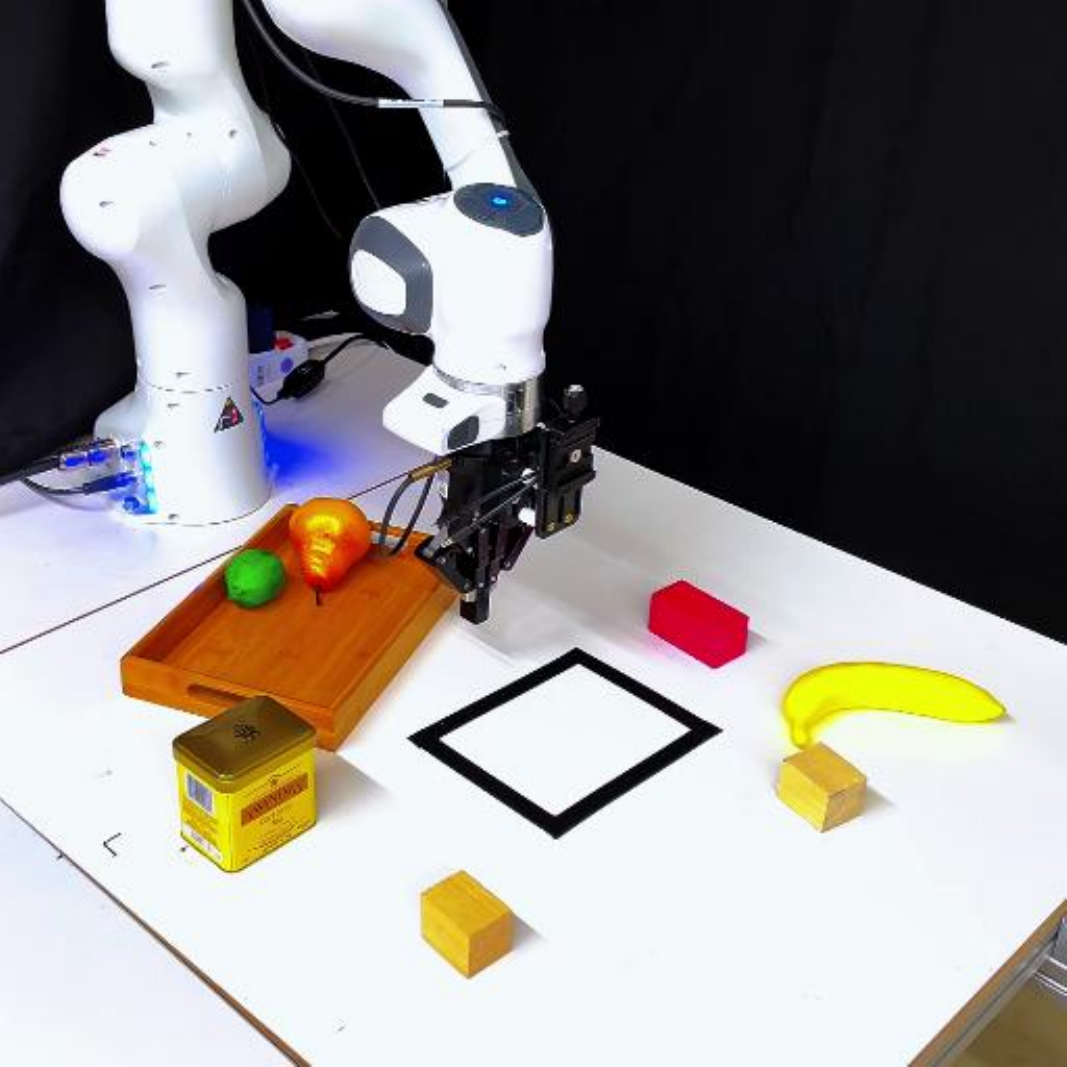}};

    \node[rectangle, rounded corners=2pt, clip, right=0.005cm of img6, ] (img7) {\includegraphics[width=0.18\textwidth]{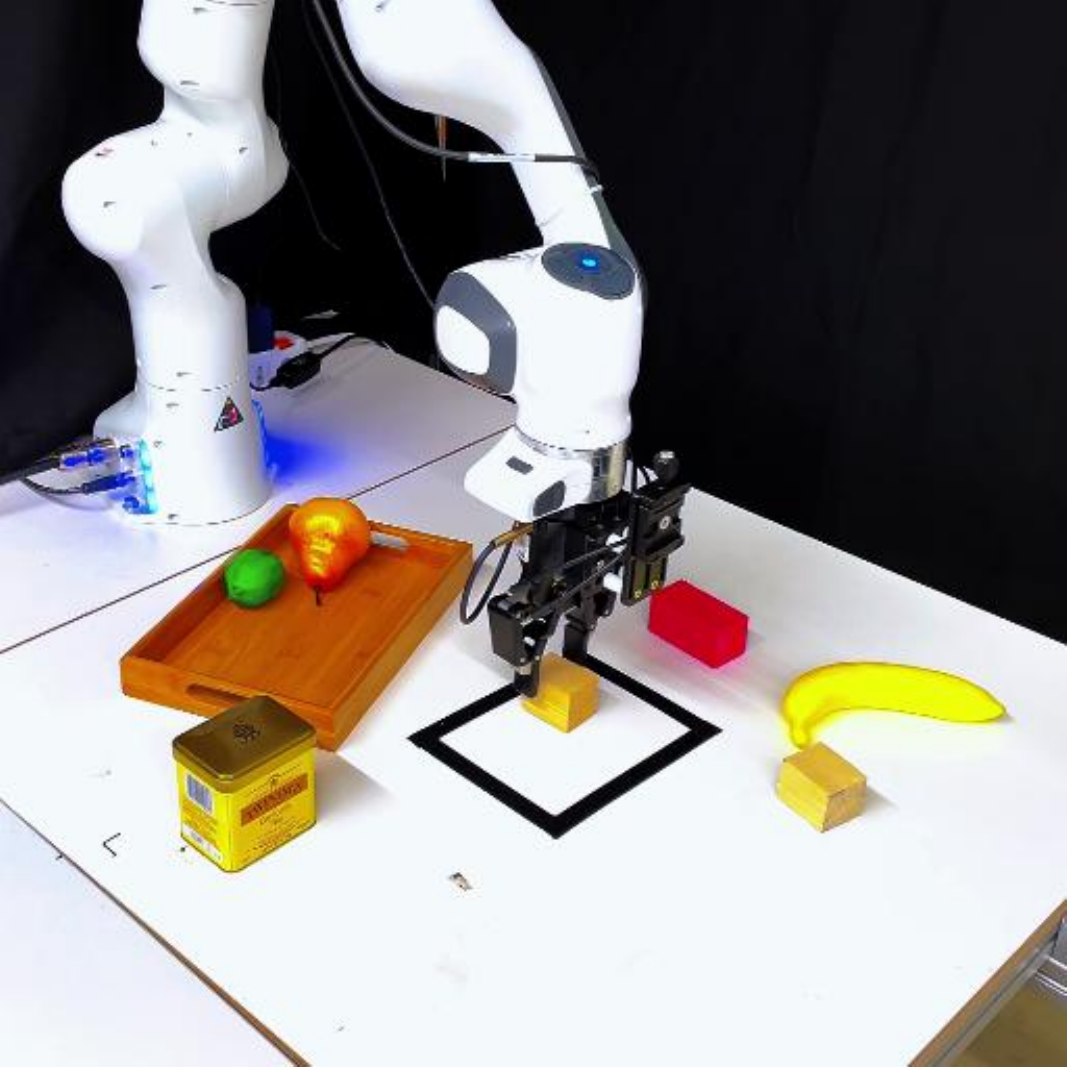}};

    \node[rectangle, rounded corners=2pt, clip, right=0.005cm of img7, ] (img8) {\includegraphics[width=0.18\textwidth]{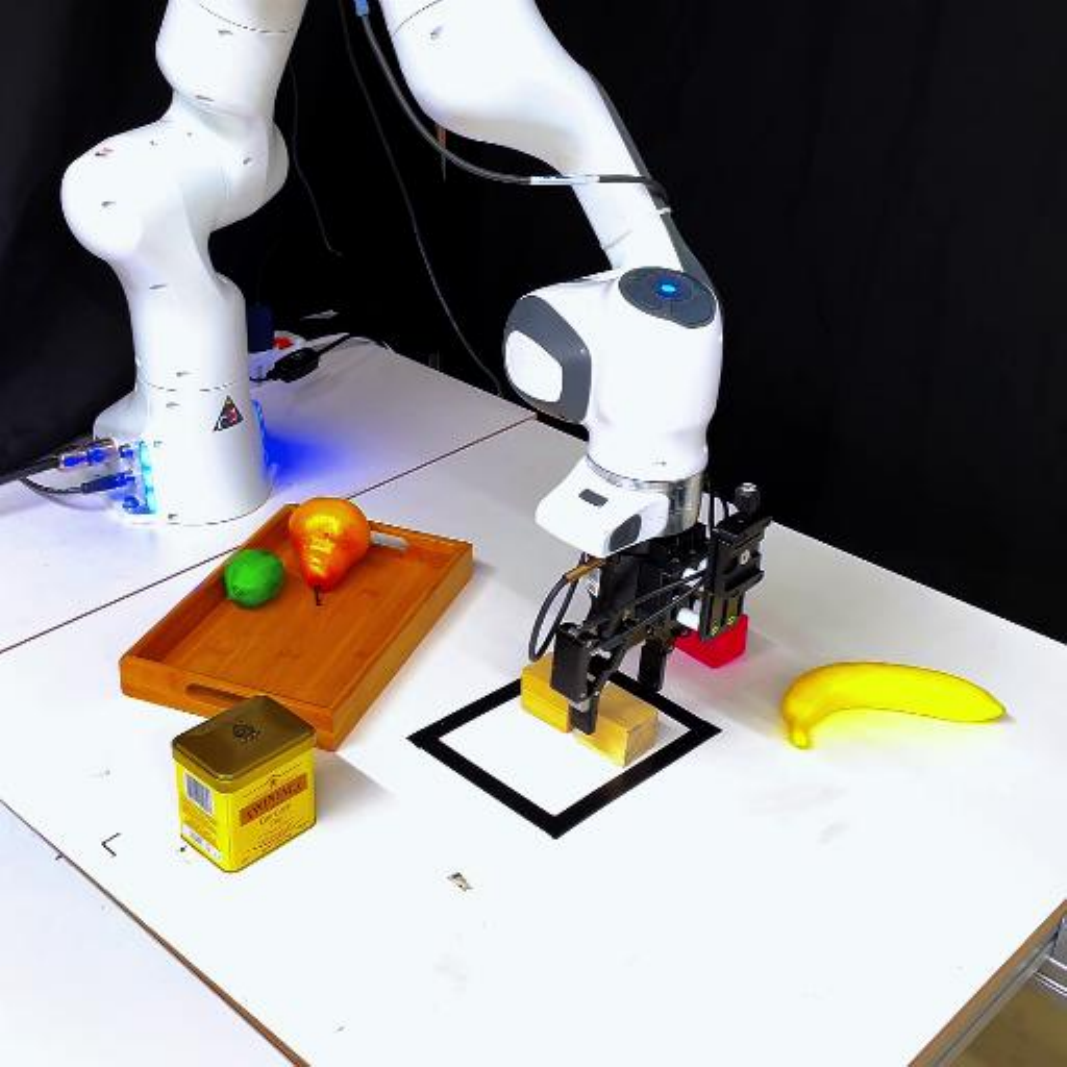}};

    \node[rectangle, rounded corners=2pt, clip, right=0.005cm of img8, ] (img9) {\includegraphics[width=0.18\textwidth]{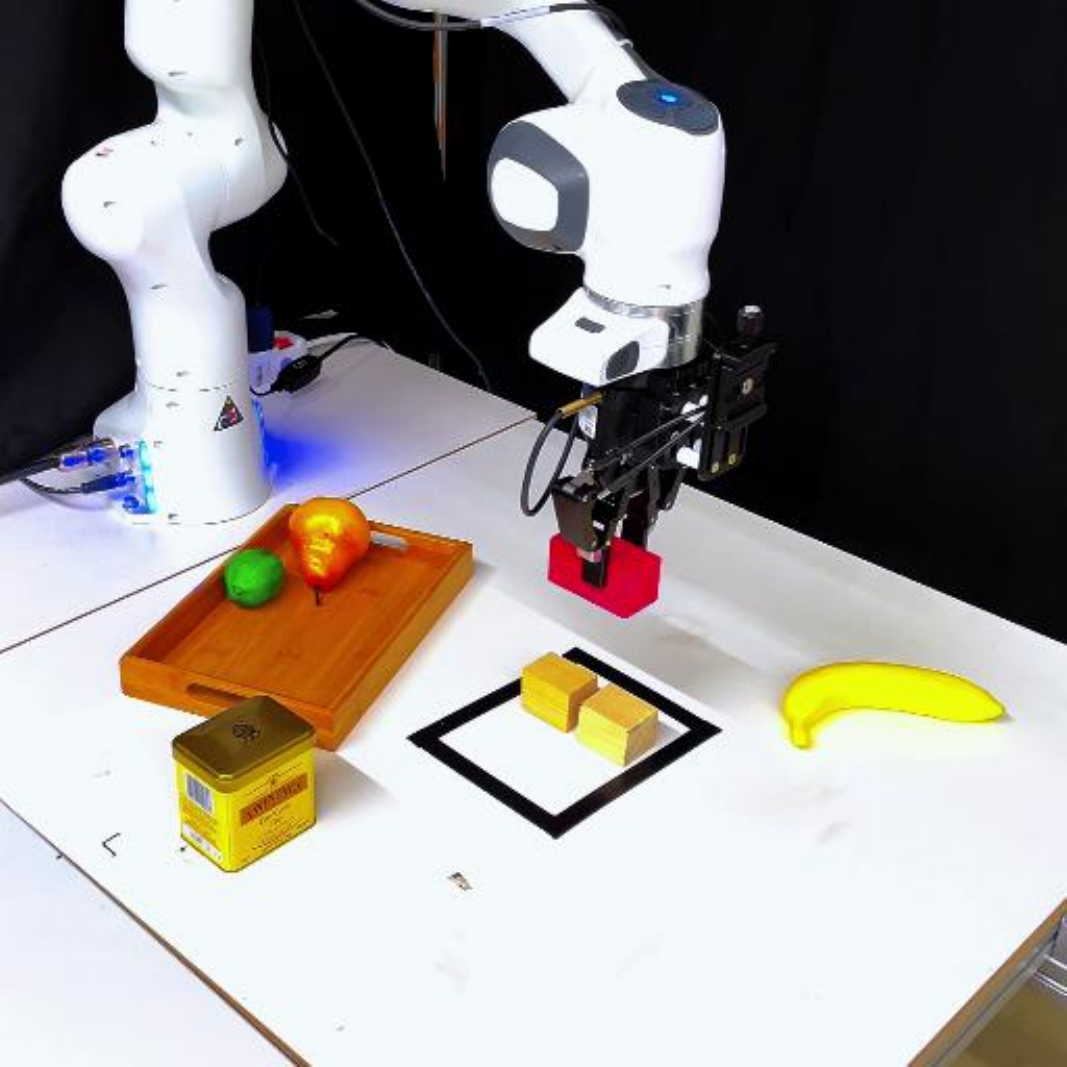}};

    \node[rectangle, rounded corners=2pt, clip, right=0.005cm of img9, ] (img10) {\includegraphics[width=0.18\textwidth]{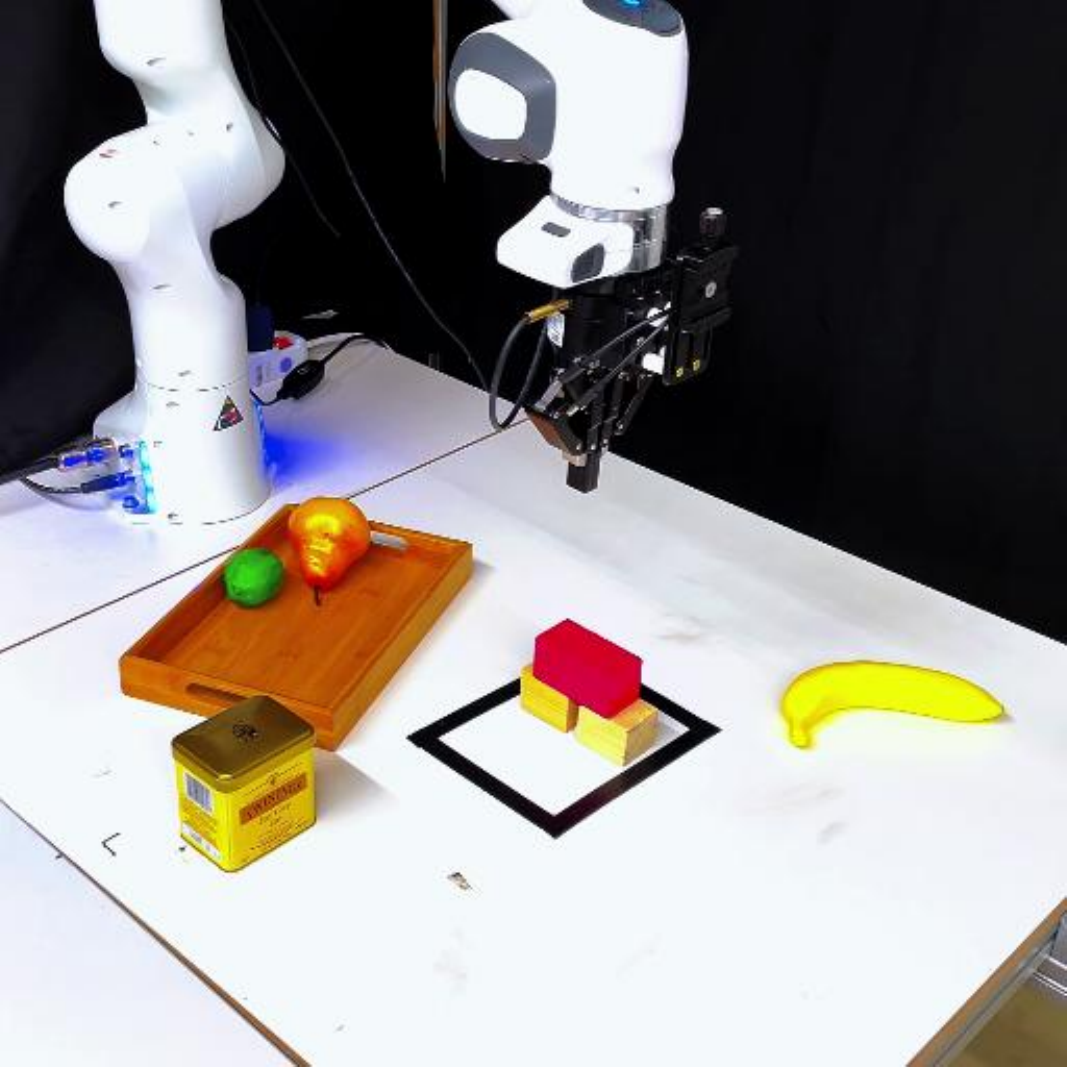}};

    \node[rectangle, clip, rounded corners=2pt, below=0.1cm of img6, ] (img11) {\includegraphics[width=0.18\textwidth]{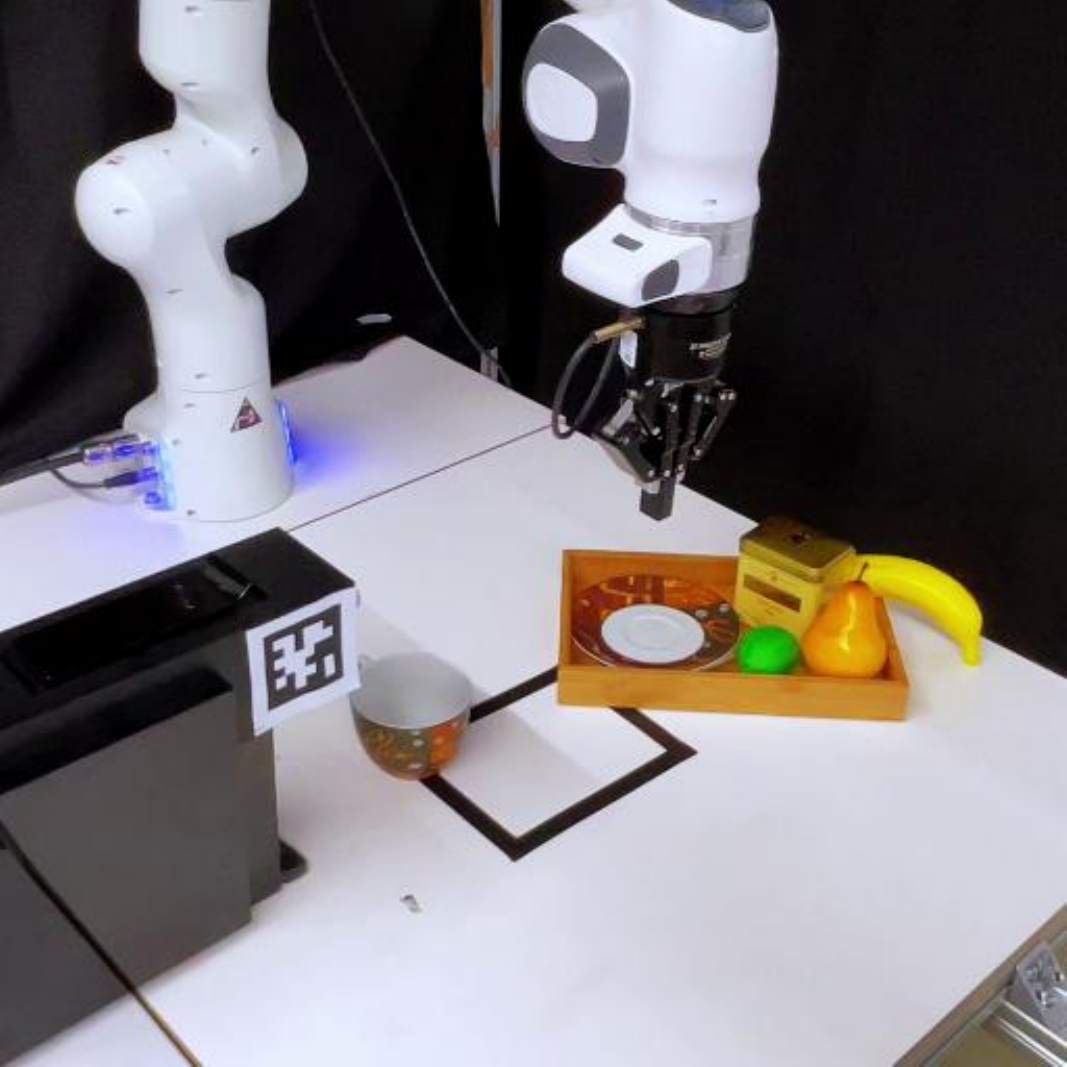}};

    \node[rectangle, rounded corners=2pt, clip, right=0.005cm of img11, ] (img12) {\includegraphics[width=0.18\textwidth]{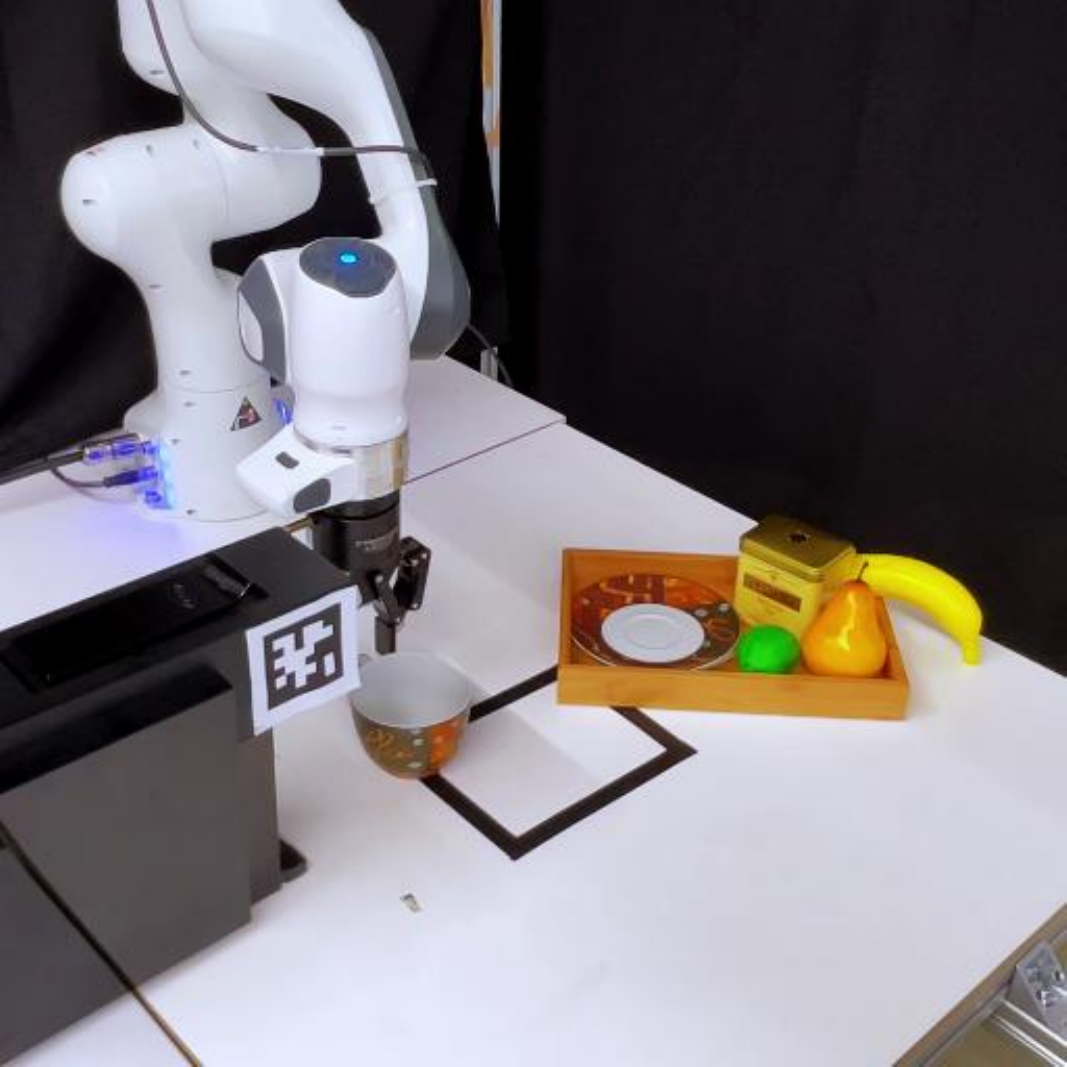}};

    \node[rectangle, rounded corners=2pt, clip, right=0.005cm of img12,] (img13) {\includegraphics[width=0.18\textwidth]{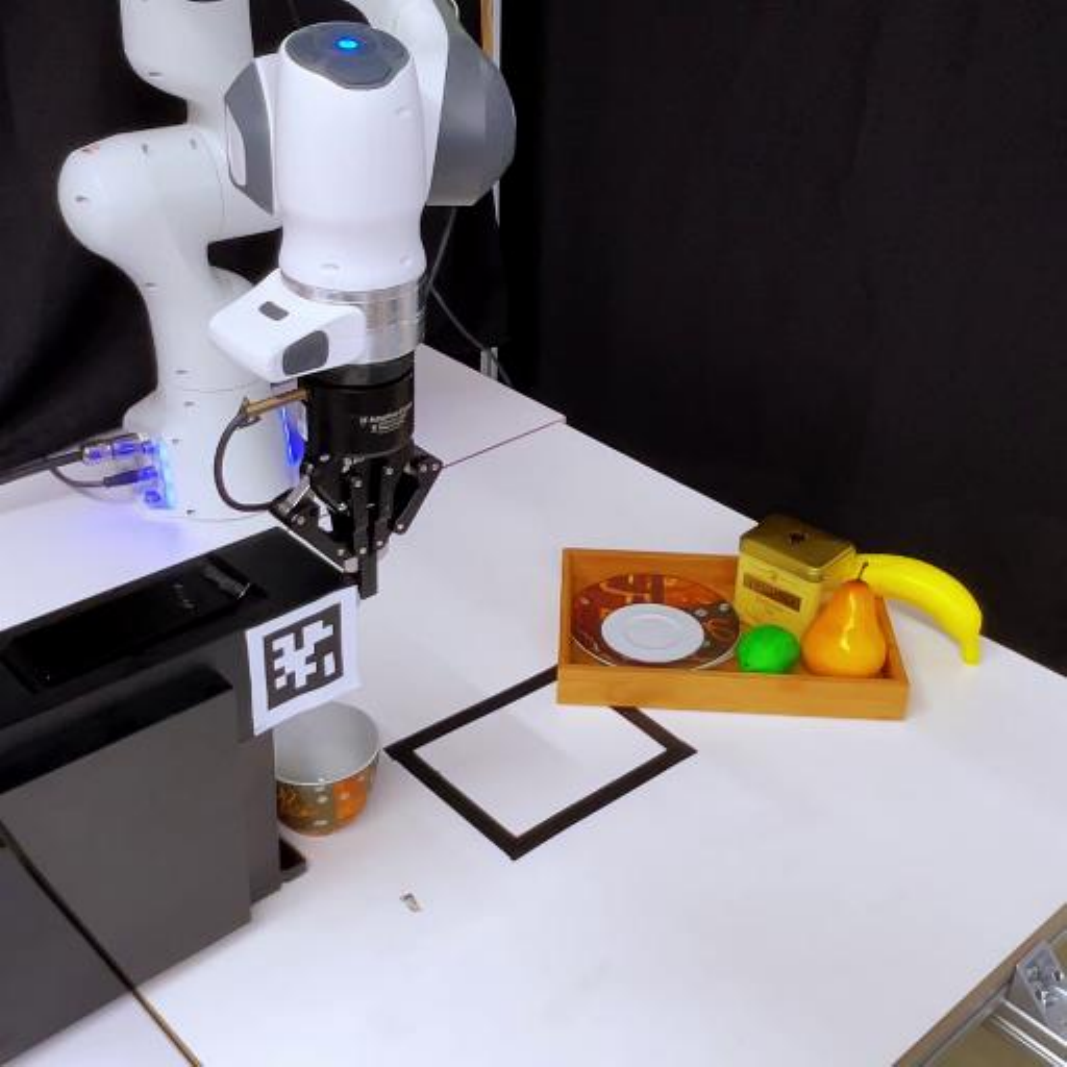}};

    \node[rectangle, rounded corners=2pt, clip, right=0.005cm of img13, ] (img14) {\includegraphics[width=0.18\textwidth]{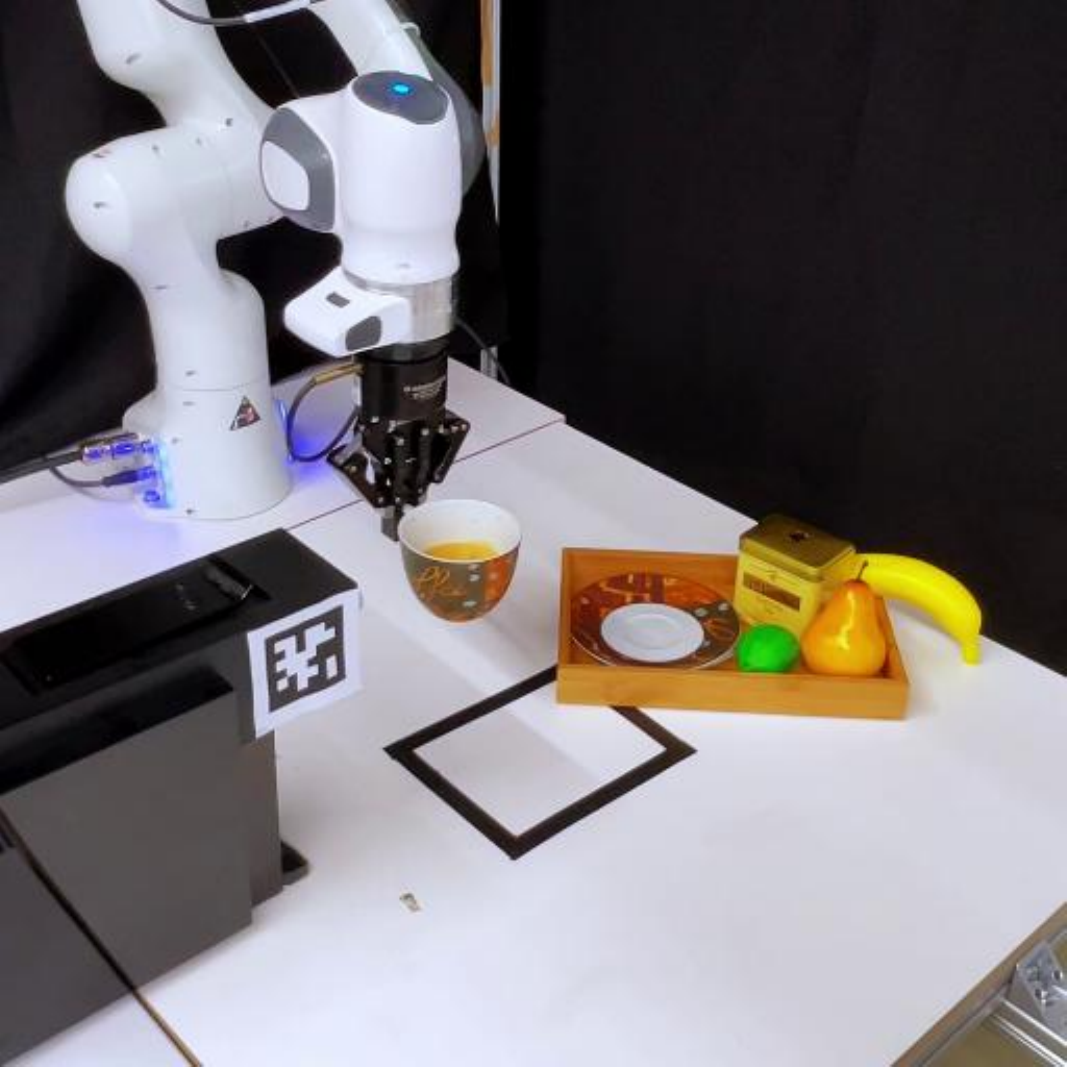}};

    \node[rectangle, rounded corners=2pt, clip, right=0.005cm of img14,] (img15) {\includegraphics[width=0.18\textwidth]{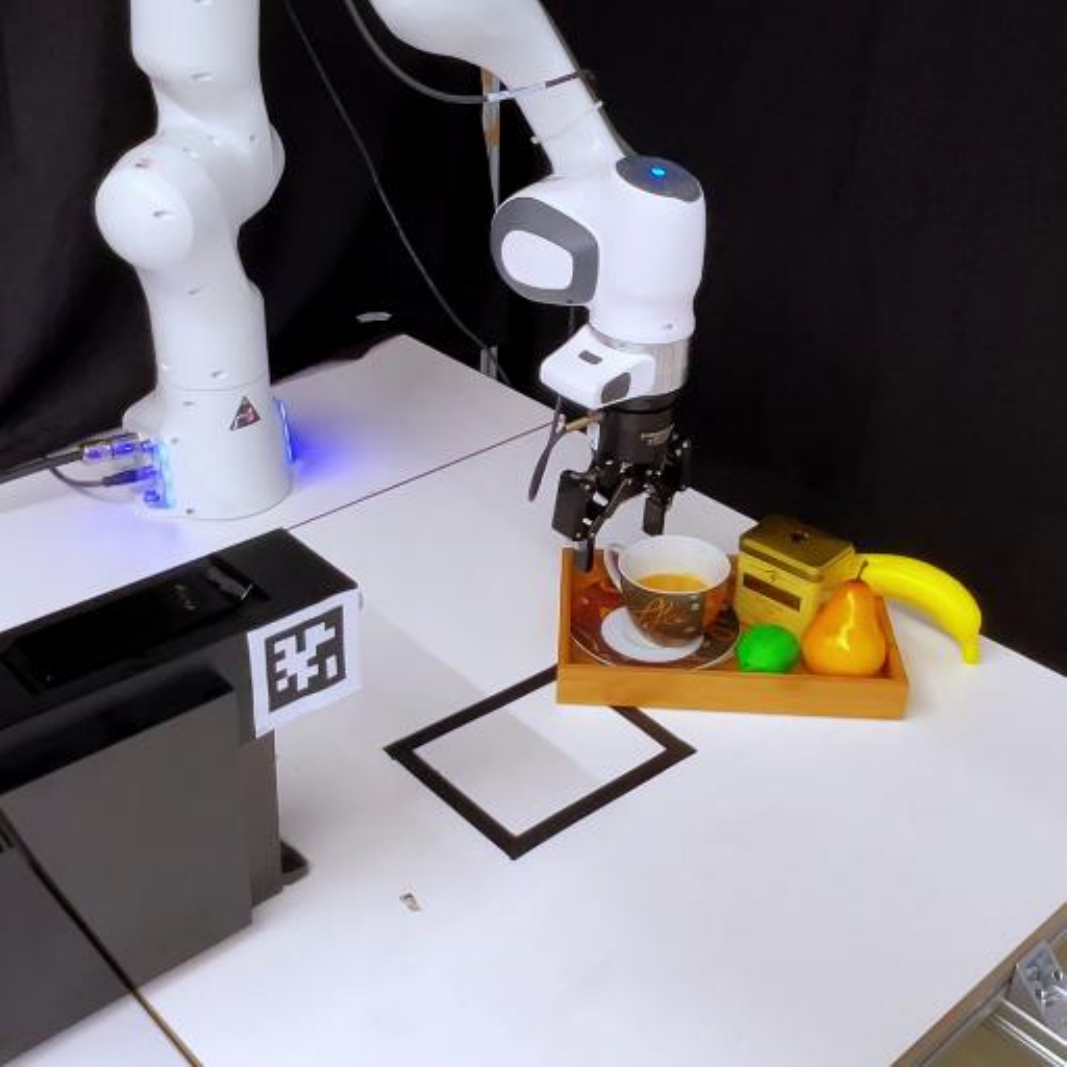}};

    \node[rectangle, clip, rounded corners=2pt, below=0.1cm of img11, ] (img16) {\includegraphics[width=0.18\textwidth]{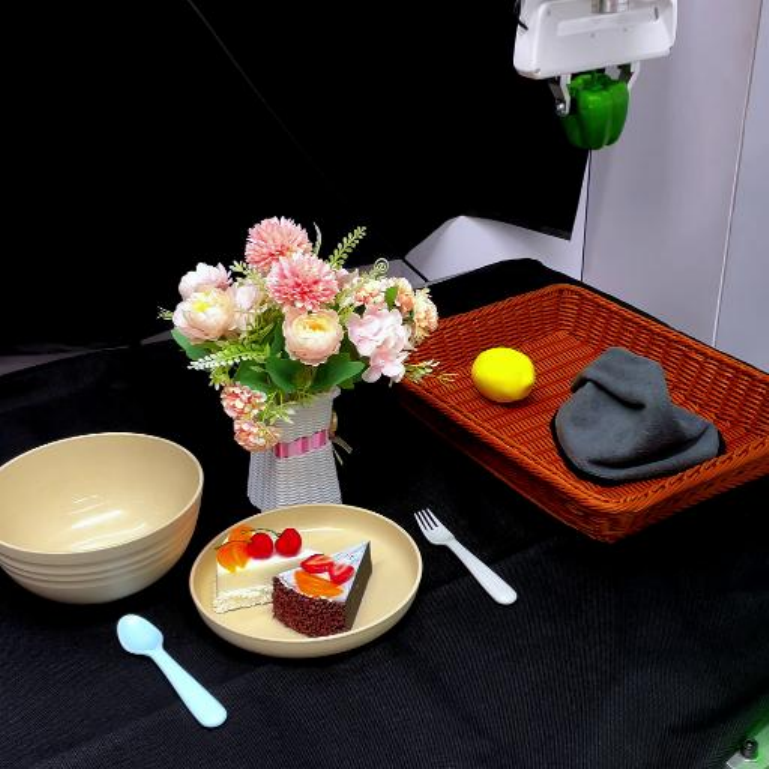}};

    \node[rectangle, rounded corners=2pt, clip, right=0.005cm of img16, ] (img17) {\includegraphics[width=0.18\textwidth]{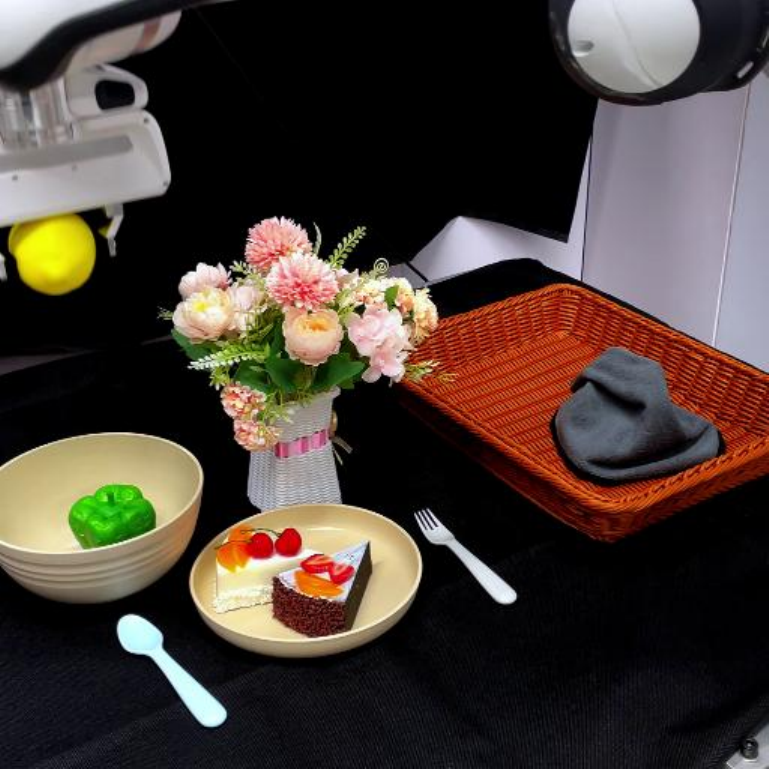}};

    \node[rectangle, rounded corners=2pt, clip, right=0.005cm of img17,] (img18) {\includegraphics[width=0.18\textwidth]{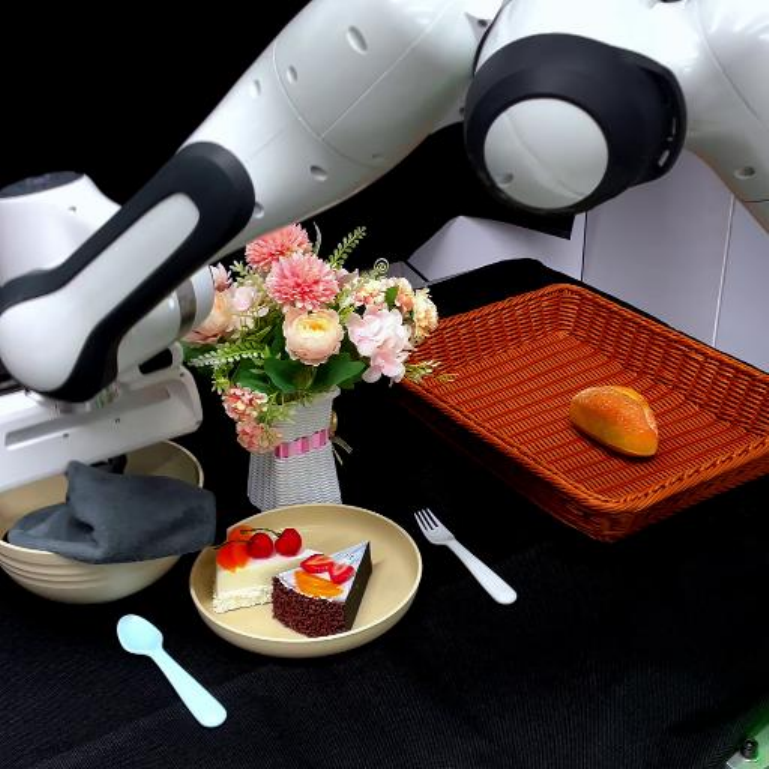}};

    \node[rectangle, rounded corners=2pt, clip, right=0.005cm of img18, ] (img19) {\includegraphics[width=0.18\textwidth]{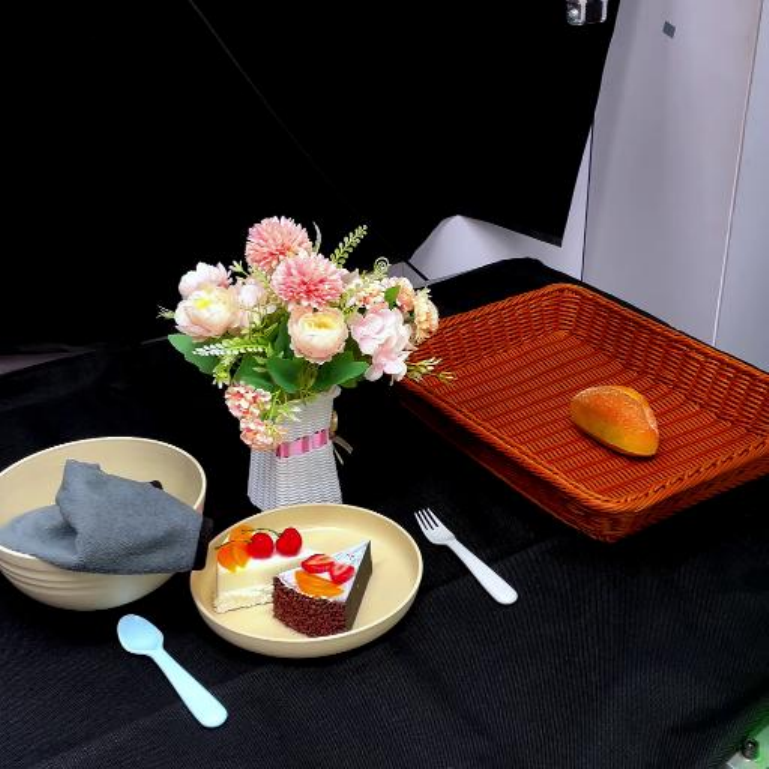}};

    \node[rectangle, rounded corners=2pt, clip, right=0.005cm of img19,] (img20) {\includegraphics[width=0.18\textwidth]{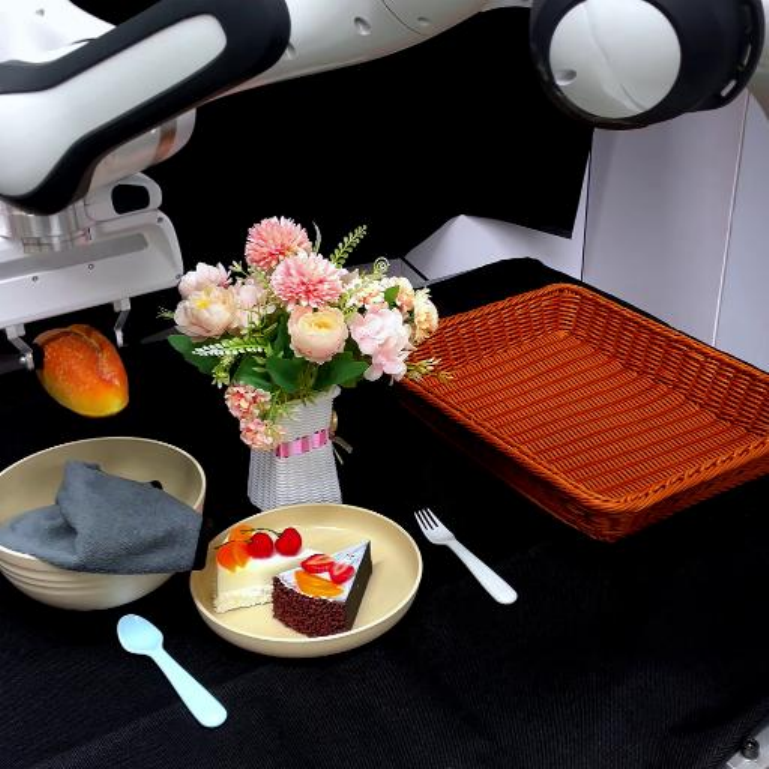}};

    \node[rectangle, draw=fforange_pv, line width=5pt, minimum width=0.18\textwidth, minimum height=0.18\textwidth]at(img19){};

    \node[rectangle, draw=ffgreen_pv!80, line width=3pt, minimum width=0.75cm, minimum height=0.5cm]at([xshift=1cm, yshift=-.175cm]img19.center){};

\end{tikzpicture}

}
\vskip -0.05in
\caption{Real-world demonstrations. Pick and place all the fruits on the plate (1st row). Stack a red head pyramid (2nd row). Make the coffee (3rd row). Detect (orange bbox) and re-pick-place occluded bread (green bbox) while avoiding the obstacle through closed-loop feedback recovery (4th row).}
\label{fig:realworld}
\vskip -0.2in
\end{figure}

%
\textbf{Real-world performance:}
Fig. \ref{fig:realworld} presents real-world demonstrations of our framework on four long-horizon tabletop manipulation tasks: 
(1) picking all fruits and placing them on a plate; (2) stacking blocks into a red head pyramid; (3) making coffee; and (4) retrieving all items from a basket, including a bread hidden under a towel, and placing them in a bowl without colliding with a nearby vase and flower.
To increase scene reasoning complexity, we introduced distractor objects such as cucumbers and pumpkins in the first scenario, requiring accurate object identification. 
In the fourth scenario, an occluded bread item challenges the system’s ability to recover from partial observations.
Before execution, the agent scans the scene using open-world foundation models, extracts object attributes and targets from user instructions, and generates task plans.
The primitives are then executed, followed by scene evaluation from the reporter, which determines task completion.
As shown in the snapshots, our framework completes the first three tasks successfully in a single loop.
In the fourth task, the hidden bread initially goes undetected. 
Upon exposure (green bbox), the reporter issues a re-planning request (orange bbox), prompting the planner with ``\textit{A bread at (x,y,z) in the basket is not yet put into the bowl.}'' to update the task plan and complete the goal. 
This highlights the framework’s robust closed-loop feedback and adaptability in real-world long-horizon scenarios. 
Videos are available in the attached files.
%

\section{Conclusions}
This paper presented DAHLIA, a data-agnostic framework for long-horizon robotic manipulation based on episodic code generation and task-level feedback. By organizing in-context examples with progressively increasing complexity and chain-of-thought reasoning, DAHLIA enables LLMs to synthesize reusable and executable task plans for complex manipulation tasks. A VLM-based reporter further provides structured outcome feedback for inter-episode replanning, improving robustness under partial observability without per-step inference. Experiments across LoHoRavens, CALVIN, Franka Kitchen, and real-world tasks demonstrate strong performance and generalization in both seen and unseen scenarios. Future work will explore adaptive example selection, stronger spatial grounding under severe occlusion, and tighter integration with open-world vision-language models.

\section*{APPENDIX}

\subsection{Prompt implementation details}\label{appendix:prompt}
For more information, refer to our code that will be open-sourced later.
The planner prompt is basically structured as below:
\begin{myverbatim}
# -----------------------------
# Existing Method Explanations
# -----------------------------
put_first_on_second(arg1 , arg2) # let robot put arg1 to arg2
get_obj_pos(obj) -> [list] # return a list of len(obj) of 3d position-vectors of obj, even when obj is just one object not a list of objects
get_obj_rot(obj) -> [list] # return a list of len(obj) of 4d quaternion orientation-vectors of obj, even when obj is just one object not a list of objects
parse_obj_name(query=str1, context=str2) -> list or str # usually return a list of str or just one str that is chosen out from str2 according to str1 which performs as description. Possible to demand for wanted return format in str1
...(For more APIs refer to our code implementation)...
# --------------------------------------
# Orientations in the Coordinate System
# --------------------------------------
left: y-    right: y+   front: x+
rear: x-    top: z+     bottom: z-
# ---------------------
# General Requirements
# ---------------------
You are writing Python code for robot manipulation, refer to the code style in the examples below. 
You can use the existing APIs above, you must NOT import other packages. Our coordinate system is 3D cartesian system, but still pay attention to the orientations. Object sizes are important in some tasks as they determine object positions, so remember to get them in advance. When you are not sure about objects or positions, you had better use parse_obj_name() and parse_position(), and clarify your return format demand. The examples below are ordered from simple to complex; use them as a curriculum to infer reusable planning patterns, and reason step-by-step (in comments) to adapt these patterns to the current task rather than copying code verbatim.
# --------------
# Task Examples
# --------------
...(examples with incremental difficulty)...
# 1. example with block placement at the target position
objects = ['yellow block with obj_id 1', 'blue bowl with obj_id 3', 'green bowl with obj_id 6']
blocks = parse_obj_name('movable yellow blocks, return result as list of object names', f'objects = {get_obj_names()}')
area = parse_position('the 2d bounding box of the top right area, return result as list of [x_min, y_min, x_max, y_max]')
for block in blocks:
    pos = get_random_free_pos(targ=None, search_area=area)
    put_first_on_second(block, pos)
# 2. example with multiple block place in a target zone.
...<code example>...
# i. example with block stack on each other.
...<code example>...
...(For more details refer to our code implementation)...
\end{myverbatim}
The prompt for the reporter has basically the same structure, except that the inputs and requirements:
\begin{myverbatim}
You are judging whether the task is completed based on the given snapshots before and after planning execution. Refer to the code examples below.
...(same instruction as planner)...
If you find the task is not complete, provide additional feedback for replanning. This should involve: objects to operate, the positions, the actions to take, and the target to achieve. 
\end{myverbatim}

\section*{ACKNOWLEDGMENT}
This work was supported by the Beijing Academy of Artificial Intelligence (BAAI).



\end{document}